\documentclass[sigconf]{acmart}

\AtBeginDocument{%
  }

\usepackage{amsmath,amsfonts}
\usepackage{algorithm}
\usepackage{algpseudocode}
\usepackage{array}
\usepackage[caption=false,font=normalsize,labelfont=sf,textfont=sf]{subfig}
\usepackage{textcomp}
\usepackage{stfloats}
\usepackage{url}
\usepackage{verbatim}
\usepackage{graphicx}
\usepackage{xcolor}
\usepackage{booktabs}
\usepackage{hyperref}
\usepackage{orcidlink}

\setcopyright{acmlicensed}
\copyrightyear{2025}
\acmYear{2025}
\acmDOI{XXXXXXX.XXXXXXX}
\acmConference[AISEC' 25]{Make sure to enter the correct
  conference title from your rights confirmation email}{October 17,
  2025}{Taipei, Taiwan}
\acmBooktitle{Proceedings of the 2025 Workshop on Artificial Intelligence and
Security (AISec ’25), October 17 2025, Taipei, Taiwan}
\acmISBN{978-1-4503-XXXX-X/2018/06}

\begin{document}

\title{Enhancing Robustness in Post-Processing Watermarking: An Ensemble Attack Network Using CNNs and Transformers}

% \affiliation{%

\author{Tzuhsuan Huang}
\authornote{Both authors contributed equally to this research.}
\affiliation{%
  \institution{Academia Sinica}
  \city{Taipei}
  \country{Taiwan}
}
\email{jason890425@citi.sinica.edu.tw}
\author{Cheng Yu Yeo}
\authornotemark[1]
\affiliation{%
  \institution{National Yang Ming Chiao Tung University}
  \city{Hsinchu}
  \country{Taiwan}
}
\email{boyyeo123.ee12@nycu.edu.tw}

\author{Tsai-Ling Huang}
\affiliation{%
  \institution{National Yang Ming Chiao Tung University}
  \city{Hsinchu}
  \country{Taiwan}
}
\email{christina.ii12@nycu.edu.tw}

\author{Hong-Han Shuai}
\affiliation{%
  \institution{National Yang Ming Chiao Tung University}
  \city{Hsinchu}
  \country{Taiwan}
}
\email{hhshuai@nycu.edu.tw}

\author{Wen-Huang Cheng}
\affiliation{%
  \institution{National Taiwan University}
  \city{Taipei}
  \country{Taiwan}
}
\email{wenhuang@csie.ntu.edu.tw}

\author{Jun-Cheng Chen}
\affiliation{%
  \institution{Academia Sinica}
  \city{Taipei}
  \country{Taiwan}
}
\email{pullpull@citi.sinica.edu.tw}

% \affiliation{%

% \affiliation{%

% \affiliation{%

% \affiliation{%

% \affiliation{%

% \affiliation{%


\begin{abstract}
Recent studies on deep watermarking have predominantly focused on in-processing watermarking, which integrates the watermarking process into image generation. However, post-processing watermarking, which embeds watermarks after image generation, offers more flexibility. It can be applied to outputs from any generative model (\textit{e.g.} GANs, diffusion models) without needing access to the model's internal structure. It also allows users to embed unique watermarks into individual images. Therefore, this study focuses on post-processing watermarking and enhances its robustness by incorporating an ensemble attack network during training. We construct various versions of attack networks using CNN and Transformer in both spatial and frequency domains to investigate how each combination influences the robustness of the watermarking model. Our results demonstrate that combining a CNN-based attack network in the spatial domain with a Transformer-based attack network in the frequency domain yields the highest robustness in watermarking models. Extensive evaluation on the WAVES benchmark, using average bit accuracy as the metric, demonstrates that our ensemble attack network significantly enhances the robustness of baseline watermarking methods under various stress tests. In particular, for the Regeneration Attack defined in WAVES, our method improves StegaStamp by 18.743\%. The code is released at \url{https://github.com/aiiu-lab/DeepRobustWatermark}.
\end{abstract}

\keywords{Deep Watermarking; Copyright Protection; Ensemble Model}

%received{20 February 2007}
%\received[revised]{12 March %2009}



\maketitle

\begin{figure}[t]
  \centering
  \includegraphics[width=1.0\linewidth]{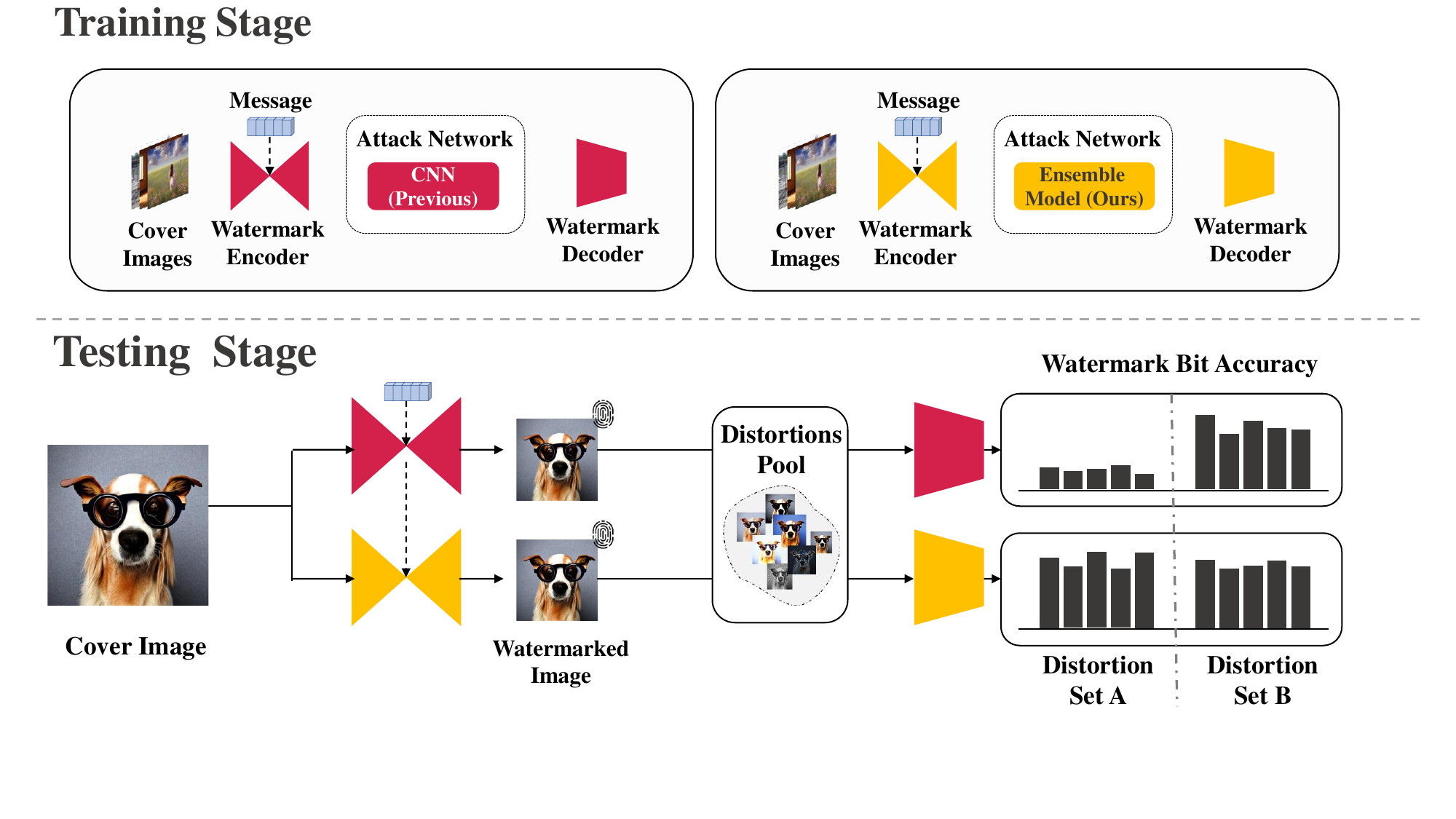}
  \vspace{-35pt}
   \caption{The \textcolor{red}{CNN-based attack network} proposed in ``Distortion Agnostic Deep Watermarking (DA)''~\cite{distortion} tends to generate limited types of distortions due to its architectural capacity, which results in suboptimal performance of the watermarking model when facing unseen attacks. To overcome this limitation, we propose an \textcolor{orange}{ensemble attack network (ensemble model)} that integrates CNN and Transformer architectures across both spatial and frequency domains to simulate a broader spectrum of attacks, thereby improving the robustness of the watermarking model. Note that the watermarking model trained with a CNN is shown in red, while the one trained with the ensemble model is shown in orange.} 
   \label{fig:teaser}
\vspace{-7pt}
\end{figure}

\section{Introduction}
\label{sec:intro}
Image generation and manipulation have advanced rapidly in recent years, mainly driven by breakthroughs in generative adversarial networks (GANs)~\cite{stylegan2, stylegan3, dual_gan, inclusive_gan, df-gan, laf, scaling, styleres, e4s, inter} and diffusion models~\cite{edm, instructpix2pix, glide, controlnet}. However, with these advances, concerns over intellectual property and copyright infringement have come to the forefront, highlighting the growing importance of robust and reliable watermarking techniques.

Previous watermarking methods can be broadly categorized into post-processing and in-processing approaches. Post-processing methods embed watermarks after image generation, whereas in-processing methods integrate watermarking directly into the image generation process. In this work, we focus on post-processing techniques, which typically consist of a watermark encoder and a decoder. The encoder embeds watermarks into images, while the decoder is responsible for extracting the embedded watermarks. To improve the resilience of watermarks against noise during transmission, Zhu~\textit{et al.}~\cite{hidden} incorporate a noise layer, also known as attack network, to simulate common distortions occurring during image transmissions and malicious attacks. Although the noise layer greatly improves the tolerance of the watermarking model against expected distortions that are simulated during training, it is insufficient when handling unforeseen distortions. To address this limitation, Luo~\textit{et al.}~\cite{distortion} propose a CNN-based attack network that generates a diverse range of image perturbations, enhancing the robustness of the watermarking model to a wider range of distortions. However, we find that complex distortions (\textit{e.g.} JPEG compression) are difficult to simulate using the CNN-based attack network, which limits the ability of the watermarking model to generalize to more challenging scenarios.

This limitation motivates us to adopt a Transformer-based architecture as the backbone of the attack network. While this approach substantially improves the robustness of the watermarking model, it also leads the image encoder to produce watermarked images with noticeable artifacts. This occurs because the watermark encoder must embed stronger watermarks to resist the intense perturbations introduced by the Transformer-based attack network. This observation encourages us to explore attack networks operating beyond the spatial domain. In contrast to spatial-domain perturbations that directly alter pixel intensities and often produce visible artifacts, frequency-domain modifications, particularly those based on the 2D Discrete Cosine Transform (DCT), operate in a more organized and perceptually coherent domain. Consequently, we integrate the DCT process with the Transformer architecture (DCT-Transformer) to apply distortions in the frequency domain. This approach preserves the quality of the encoded images while enhancing the model’s resilience against a wider variety of distortions.

Moreover, we find that CNN-based and DCT-Transformer-based attack networks simulate distinct types of distortion, which motivates us to employ an ensemble technique to construct attack networks in different configurations (\textit{e.g.} model cascade or parallel as described in Section~\ref{sec:ensemble}). Our experiments demonstrate that utilizing an ensemble model can harness the advantages of both CNN and Transformer under different domains, as shown in Figure~\ref{fig:teaser}, enhancing robustness of watermarks under several distortions.

To demonstrate the efficacy of the proposed ensemble attack network, we train several watermarking models using our attack network and evaluate their robustness across a wide range of attacks. Following the WAVES benchmark~\cite{waves}, we apply three attacks (distortion, embedding, and regeneration) defined in WAVES to evaluate the robustness of the watermarking models. In addition, we utilize off-the-shelf image editing models to attack (manipulate) watermarked images and assess whether the watermarks remain after editing. \textbf{(1) Distortion Attacks (WAVES):} We assess the resilience of the watermarks by applying a comprehensive set of distortions that simulate realistic transmission and image processing scenarios. By integrating our ensemble attack network, we significantly enhance the robustness of baseline watermarking methods, including HiDDeN~\cite{hidden}, StegaStamp~\cite{stegastamp}, and Stable Signature~\cite{signature}, achieving average bit accuracy improvements of 6.995\%, 5.395\%, and 8.386\%, respectively. \textbf{(2) Embedding Attacks (WAVES):} Against embedding attacks that utilize off-the-shelf embedding models to fool the watermark decoder, our method achieves nearly 100\% bit accuracy on HiDDeN and StegaStamp, and 90.078\% on Stable Signature. \textbf{(3) Regeneration Attacks (WAVES):} In regeneration attacks, we employ diffusion models or VAEs to alter the image’s latent representation. Our approach outperforms ``Distortion Agnostic Deep Watermarking (DA)'' and StegaStamp by 3.537\% and 18.743\%, respectively, while achieving performance comparable to Stable Signature. All experiments mentioned above are conducted on the COCO dataset. For \textbf{(4) Manipulation Attacks}: Our method defeats DA and StegaStamp by 9.963\% and 7.551\% in average bit accuracy on the COCO and CelebA datasets.

Our main contributions are twofold: (1) We present a comprehensive study on the robustness of watermarking models trained with different attack networks. Specifically, we propose using Transformer architectures as alternatives to CNNs for simulating adversarial distortions and observe that they generate distinct types of distortions. Moreover, we introduce frequency-domain perturbations by incorporating DCT processing into both CNN- and Transformer-based architectures, providing a complementary perspective to spatial-domain attacks.
(2) We propose a novel ensemble attack network that combines a CNN-based attack network operating in the spatial domain with a Transformer-based attack network operating in the frequency domain. We demonstrate that this ensemble technique significantly improves the robustness of several baseline watermarking methods, as evaluated on the WAVES benchmark.

\begin{figure*}[t]
  \centering
  \includegraphics[width=0.9\linewidth]{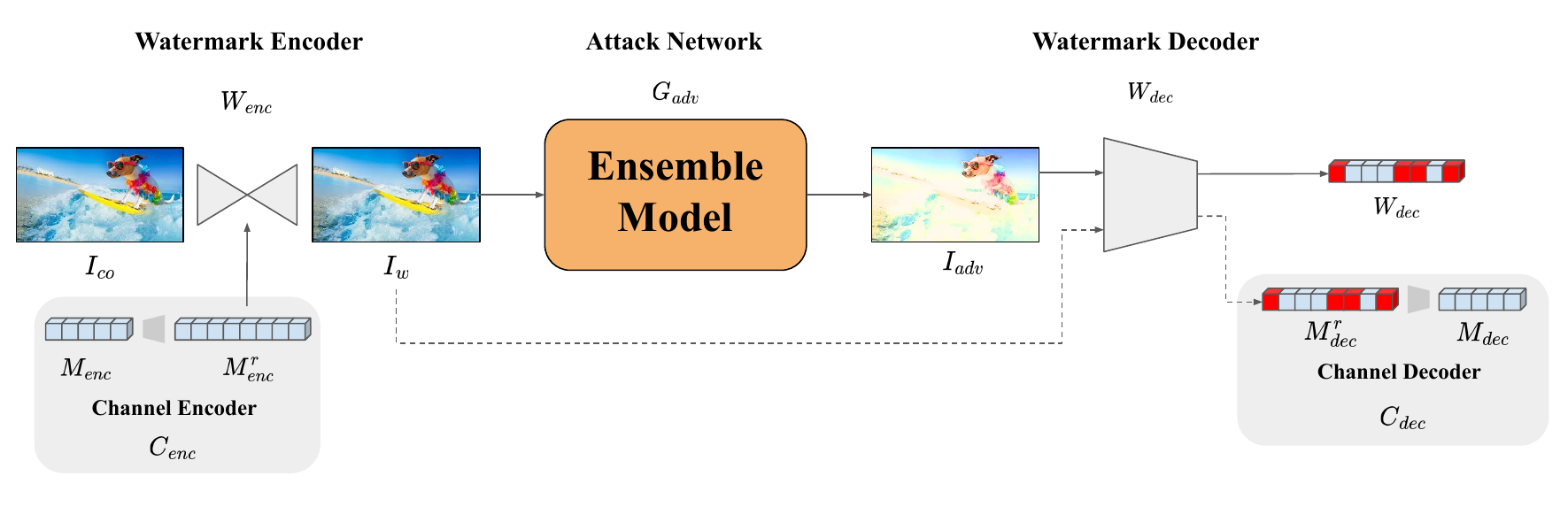}
  \vspace{-17pt}
  \caption{Our framework begins with the channel encoder $C_{enc}$, which injects redundancy into the encoded message $M_{enc}$ and produces a redundant message $M^{r}{enc}$. The watermark encoder $W{enc}$ then embeds $M^{r}{enc}$ into the cover image $I_{co}$ to create the watermarked image $I_{w}$. The attack network $G_{adv}$ generates an adversarial image $I_{adv}$ from $I_{w}$, simulating potential transmission or processing distortions. The watermark decoder $W_{dec}$ extracts redundant decoded messages $M^{r}{dec}$ and $M^{r}{adv}$ from $I_{w}$ and $I_{adv}$, which are then sent to the channel decoder $C_{dec}$ to convert them back to the decoded message $M_{dec}$. 
  }
  \label{fig:overview}
\end{figure*}
\section{Related Work}
\label{sec:related_works}
% Recently, deep watermarking~\cite{two_stage, exploring_watermark, stegastamp, hidden, distortion, de-end, estimating_watermark,chen2025invisible} has gained significant attention due to its ability to protect copyrights from various attacks. 
In this section, we briefly review the recent developments in deep learning-based image watermarking approaches~\cite{two_stage, exploring_watermark, stegastamp, hidden, distortion, de-end, estimating_watermark,chen2025invisible}. Following WAVES benchmark~\cite{waves}, current research on deep watermarking can be divided into two categories: (1) post-processing watermarking and (2) in-processing watermarking.

\subsection{Post-Processing Watermarking} In general, post-processing watermarking employs deep learning techniques to embed information into generated or existing images using an encoder–decoder framework. The encoder embeds a sequence of binary messages or secret images (watermarks) into a cover image, allowing the decoder to recover the embedded message even if the watermarked image is degraded. HiDDeN~\cite{hidden} was the first method to adopt this end-to-end framework for watermarking. To handle distortions introduced during digital transmission, Zhu~\textit{et al.}~\cite{hidden} introduce a noise layer to simulate a range of transformations during training, forcing the model to learn watermark encodings that can survive such distortions. However, this approach has a significant limitation, as it requires the noise layer to be differentiable. In practice, many common distortion operations are non-differentiable, making it impractical to incorporate them within the noise layer. To address these limitations, Fang~\textit{et al.}~\cite{watermark_enhance_reala_scene} propose a mask-guided frequency enhancement algorithm to improve the watermarking network, enhancing its robustness against practical distortions. Inheriting the concept of HiDDeN, Luo~\textit{et al.}~\cite{distortion} replace the noise layer with the CNN-based attack network with adversarial training to cover both known and unseen distortions. They also leverage channel coding to further enhance the robustness of watermarks. In StegaStamp~\cite{stegastamp}, Tancik~\textit{et al.} propose an encoder-decoder framework that enables hyperlinks embedded in images to remain intact even after physical transmission. Although StegaStamp is highly robust to distortions that occur during image transmission in the wild, it has a relatively low capacity for embedding information. To break this restriction, Lu~\textit{et al.}~\cite{dwt-cnn} adopt DWT transform and Inverse DWT (IDWT) transform as down-sampling and up-sampling layers used in watermarking process to improve the performance of StegaStamp. To achieve watermark robustness against print-camera (P-C) noise, Qin~\textit{et al.}~\cite{print_camera} designed a deep noise simulation network to simulate the fusion of real P-C noises, enhancing the robustness of the watermark.

\subsection{In-Processing Watermarking} In-processing watermarking integrates watermark embedding into the image generation process. ``Artificial Fingerprinting for Generative Models (AF)''~\cite{af} is the first work to investigate the transferability of watermarks in generative models. Yu~\textit{et al.}~\cite{af} embed watermarks into the training data and then use the watermarked data to train GANs, assessing whether GAN can generate images that retain the embedded watermarks. They demonstrate that their watermarking solution can be applied to a variety of generative models and ensure the robustness of the watermarks. However, the robustness of the watermarks is insufficient to withstand certain attacks. Recent approaches have focused on integrating watermark embedding into diffusion models (DMs), further enhancing watermark robustness. In ``A Recipe for Watermarking Diffusion Models (Recipe)''~\cite{recipe}, Zhao~\textit{et al.} undertake a thorough investigation for customizing traditional watermarking methods to effectively integrate watermarking process into cutting-edge DMs, such as Stable Diffusion~\cite{ldm}. Fernandez~\textit{et al.}~\cite{signature} fine-tune the VAE decoder within the latent diffusion models to generate images containing predefined watermarks. Wen~\textit{et al.}~\cite{tree-ring} embed the watermarks into the latent noise of DMs under frequency domain, making watermarks more resistant to removal by malicious users. Lukas~\textit{et al.}~\cite{ptw} fine-tune the generative model, using model's prior knowledge and a well-trained watermark decoder to produce watermarked images. Kim~\textit{et al.}~\cite{wouaf} introduce a weight modulation process, enabling the model to embed various watermarks into generated images rather than being limited to a single predefined one. 

In our experiments, we demonstrate that the proposed method improves the robustness of post-processing watermarking models. We further show that it also enhances the robustness of Stable Signature~\cite{signature}, an in-processing watermarking method.

\begin{table*}[h]
\caption{We calculate the bit accuracy between encoded and decoded watermarks, evaluating them under various distortions as defined in WAVES~\cite{waves}. ``Methods'' and ``Trans'' indicate different attack networks and Transformer, respectively. The CNN mentioned here is our re-implementation of ``Distortion Agnostic Deep Watermarking (DA)''~\cite{distortion}, as the authors, Luo~\textit{et al.}~\cite{distortion}, did not release their code and pre-trained weights. All networks are trained and evaluated on the COCO dataset. The parameter $p$ represents the strength of the distortion; for instance, in the resizedcrop attack, 10\% to 15\% of the watermarked image is cropped before being sent to the decoder to extract the watermarks. The highest score is highlighted in \textbf{bold}. Note that the CNN performs best on red-marked distortions, while the DCT-Transformer performs best on blue-marked distortions.}
\label{tab:analysis_network_coco}
\vspace{-5pt}
\scalebox{0.78}{
\begin{tabular}{cl|ccccccccccccc|c}
\toprule
& Methods & Identity & \begin{tabular}[c]{@{}c@{}}\textcolor{red}{resizedcrop}\\ $p=10 \sim 15$\%\end{tabular} & \begin{tabular}[c]{@{}c@{}}\textcolor{red}{erasing}\\ $p=5 \sim 25\%$\end{tabular} & \begin{tabular}[c]{@{}c@{}}\textcolor{red}{brightness}\\ $p=20 \sim 100\%$\end{tabular} & \begin{tabular}[c]{@{}c@{}}\textcolor{black}{blurring}\\ $p=4 \sim 20 pix$\end{tabular} & \begin{tabular}[c]{@{}c@{}}\textcolor{black}{rotation}\\ $p=9^\circ \sim 45^\circ$ \end{tabular} & \begin{tabular}[c]{@{}c@{}}\textcolor{blue}{contrast}\\ $p=20 \sim 100\%$\end{tabular} & \begin{tabular}[c]{@{}c@{}}\textcolor{blue}{noise}\\ $std=0.02 \sim 0.1$\end{tabular} & \begin{tabular}[c]{@{}c@{}}\textcolor{blue}{compression}\\ $p=90 \sim 10$\end{tabular} & Avg \\
\hline
(a) & CNN~\cite{distortion} & 99.947 & \textbf{90.301} & \textbf{99.1} & \textbf{97.917} & 49.991 & 52.699 & 97.852 & 58.793 & 68.423 & 79.447 \\ 
& Trans & 97.472 & 64.596 & 95.383 & 92.018 & \textbf{53.038} & \textbf{59.935} & 92.786 & 91.348 & 90.828 & 81.933 \\
\midrule
(b) & DCT-CNN & 99.294 & 74.366 & 96.579 & 94.563 & 50.141 & 54.508 & 95.003 & 77.131 & 79.014 & 80.066 \\
& DCT-Trans & 99.678 & 86.824 & 93.965 & 97.799 & 49.909 & 58.046 & \textbf{98.234} & \textbf{95.772} & \textbf{93.007} & \textbf{85.914} \\
\bottomrule
\end{tabular}
}
\end{table*}
\section{The Proposed Method}
\label{sec:analysis}
In Section~\ref{sec:pre}, we introduce the previous watermarking pipeline, including the overall framework and the objective loss used to train the model. Section~\ref{sec:cnn_vs_transformer} highlights the limitations of the attack network (CNN) employed in ``Distortion Agnostic Deep Watermarking (DA)''~\cite{distortion} and presents a new Transformer-based alternative. In Section~\ref{sec:dct_vs_spatial}, we assess the effectiveness of different architectures in both the spatial and frequency domains, providing detailed explanations and results. Finally, Section~\ref{sec:ensemble} explores various ensemble techniques aimed at combining the strengths of different attack networks under multiple configurations.

\subsection{Preliminaries}
\label{sec:pre}
To evaluate the efficacy of our attack network, we integrate it into two state-of-the-art (SOTA) post-processing watermarking methods: HiDDeN~\cite{hidden} and StegaStamp~\cite{stegastamp}. Following DA~\cite{distortion}, we incorporate channel coding (NECST)~\cite{necst} into the watermarking pipeline of HiDDeN. As illustrated in Figure~\ref{fig:overview}, the pipeline consists of three components: a channel coding model, a watermarking model, and an attack network. The channel coding model is composed of a channel encoder $C_{enc}$ and a decoder $C_{dec}$. The watermarking model is composed of a watermarking encoder $W_{enc}$ and a watermarking decoder $W_{dec}$. The attack network $G_{adv}$ in HiDDeN is composed of several pre-defined noises including Crop, Gaussian Blur, etc. The training pipeline can be divided into the following three parts.

\noindent\textbf{Channel Coding.} The objective loss used to train channel coding model is shown below
\begin{equation}
{\footnotesize
\scalebox{1.0}{
    \label{eqn:channel_enc_loss}
    \begin{math}
    \begin{aligned}
        \mathcal{L}_C(\boldsymbol\theta_{\text{NECST}}) = -\frac{1}{L_\text{M}}\sum_{i=0}^{L_\text{M}}M_i \cdot \text{log}(\sigma(\hat{M_i})) + (1-M_i) \cdot \text{log}(1-\sigma(\hat{M_i})),
\end{aligned}
    \end{math}
 }
}
\end{equation}
where $\boldsymbol{\theta}_{\text{NECST}}$ denotes the parameters of the channel coding model, $L_\text{M}$ is the length of the original binary message $M_i$, and $\hat{M_i}$ is the recovered message from the NECST model. The function $\sigma$ represents the sigmoid function. Note that the NECST model is optimized independently of the watermarking model.

\noindent\textbf{Watermarking Models.} The watermark encoder $W_{enc}$ is optimized using the objective loss shown below
\begin{equation}
{
\scalebox{1.0}{
    \label{eqn:watermark_encoder_loss}
    \begin{math}
    \begin{aligned}
           \mathcal{L}_E(\boldsymbol\theta_{W_{enc}}) = \alpha_{W_{enc}}^1 \| I_{co} - I_w \|^2 + \alpha_{W_{enc}}^2 \mathcal{L}_G (I_{w}),
    \end{aligned}
    \end{math}
 }
}
\end{equation}
where $\alpha_{W_{enc}}^1$, $\alpha_{W_{enc}}^2$ are hyperparameters for the loss function, and $I_{co}$, $I_{w}$ refer to the cover image and the watermarked image, respectively. The GAN loss $\mathcal{L}_G$ is employed to enhance the quality of watermarked images by jointly training a discriminator, which aims to minimize the perceptible differences between the cover and watermarked images. This ensures that the modifications introduced during the watermarking process are virtually undetectable. 

The watermarking decoder $W_{dec}$ is optimized using the objective loss shown below
\begin{equation}
{
\scalebox{0.9}{
    \label{eqn:watermark_decoder_loss}
    \begin{math}
    \begin{aligned}
            \mathcal{L}_D(\boldsymbol\theta_{W_{dec}}) = \alpha_{W_{dec}}^1 \| M_{dec}^r - M_{enc}^r \|^2 + \alpha_{W_{dec}}^2 \| M_{adv}^r - M_{enc}^r \|^2,
    \end{aligned}
    \end{math}
 }
}
\end{equation}
where $\alpha_{W_{dec}}^1$, $\alpha_{W_{dec}}^2$ are hyperparameters for the loss function, and $M_{dec}^r$, $M_{enc}^r$, $M_{adv}^r$ refer to the redundant decoded messages, redundant encoded messages (watermarks after channel coding), and redundant adversarial decoded messages, respectively. The watermark decoder is optimized by minimizing the mean squared error between the original messages (watermarks) and the decoded messages from both watermarked and adversarial images.

\noindent\textbf{Attack Network.} The attack network $G_{adv}$ aims to generate distortions that can enhance the robustness of the watermarking model. The objective loss for the attack network is shown below
\begin{equation}
{
\scalebox{1.0}{
    \label{eqn:attack_network}
    \begin{math}
    \begin{aligned}
   \mathcal{L}_{adv}(\boldsymbol\theta_{G_{adv}})  =\alpha_{adv}^1 \| I_{adv} - I_{w} \|^2 - \alpha_{adv}^2 \| M_{adv}^{r} - M_{enc}^{r} \|^2,
    \end{aligned}
    \end{math}
 }
}
\end{equation}
where $\alpha_{adv}^1$, $\alpha_{adv}^2$ are hyperparameters for the loss function, and $I_{adv}$, $I_{w}$ refer to adversarial images and watermarked images, respectively. The coefficient $\alpha_{adv}^1$ encourages the minimization of differences between the watermarked and adversarial images, while $\alpha_{adv}^2$ guides the attack network to generate adversarial examples by injecting noise into the watermarked images.

In StegaStamp, the training pipeline follows a similar structure, with the key difference being the absence of a channel coding.

In this work, we focus on analyzing the influence of different attack networks on the robustness of watermarking models. The comparison of different architectures and domains is provided in the following sections.

\subsection{Analysis of CNN and Transformer}
\label{sec:cnn_vs_transformer}
Although it has been demonstrated that a CNN-based attack network~\cite{distortion} can effectively improve the performance of the watermarking models through adversarial training, there is still an imperfection in robustness for certain complex distortions (\textit{e.g.} JPEG compression). Our analysis indicates that the limitation of the CNN architecture stems from its restricted kernel size and receptive field. Despite this limitation can be alleviated by stacking more layers into the CNN model to expand the receptive field, Luo et al.~\cite{distortion} mention that an increase in model complexity can lead to a decrease in the robustness of watermarks. To solve this problem, we investigate a Transformer architecture as a potential solution. To employ the Transformer as the model backbone in adversarial training, the watermarked image is first divided into $8\times8$ image patches and projected into patch embedding tokens. These tokens are then processed by the Transformer via the multi-head self-attention:
\begin{equation}
{
\scalebox{0.8}{
    \label{eqn:multihead}
    \begin{math}
    \begin{aligned}
         \textbf{Multihead} (\boldsymbol{X_{e}}) = \textbf{Concat}_{i=1}^{m} \Biggl\{  \text{softmax}(\frac{W_{i}^{Q}X_{e}(W_{i}^{K}X_{e})^T}{\sqrt{d_{K}}}) W_{i}^{V}X_{e} \Biggl\},
    \end{aligned}
    \end{math}
 }
}
\end{equation}
where $d_k$ is the dimensionality of the key vectors and \textbf{Concat} denotes the concatenation operation with $m$ heads, $X_e$ represents the input embedding tokens, and $W^Q_i, W^K_i, W^V_{i}$ are trainable matrices of weight parameters that transform features from $X_e$ to get the query, key, and value matrices respectively. 

Although the Transformer-based attack network improves watermark robustness, as shown in Table~\ref{tab:analysis_network_coco}(a), it also reduces the \textit{Identity value}, which measures decoding ability on clean (undistorted) watermarked images. A low Identity value means the decoder cannot reliably extract the watermark even without any distortions, making the model impractical. This suggests that while strong spatial-domain distortions enhance robustness against many attacks, they can also interfere with the model’s basic decoding ability. To mitigate this issue, we investigate attack networks that operate in the frequency domain, which can simulate challenging distortions while better preserving the Identity value. \\
\vspace{-10pt}

\subsection{Analysis of Spatial and Frequency Domains}
\label{sec:dct_vs_spatial}
First of all, we argue that the adversarial loss function~\eqref{eqn:attack_network} for the attack network is incompatible, as it is difficult to inject perturbations in the spatial domain while maintaining the quality of watermarked images, which in turn reduces the accuracy of recovering the embedded watermark from encoded images (Identity value). Our experiment shows that if the adversarial images are highly distorted due to the second term in Eq.~\eqref{eqn:attack_network}, it is hard for the watermarking model to converge during training. As shown in Table~\ref{tab:alpha_2}, the Identity value decreases significantly as $\alpha_{adv}^2$ increases. This suggests that to facilitate the convergence of the watermarking model, reducing the strength of the distortion is necessary. However, such reduction in distortion strength inevitably leads to a decrease in the robustness of watermarks against distortions. This inspires us to investigate the attack network's ability to generate adversarial images in the frequency domain. 

Unlike spatial-domain perturbations that directly alter pixel intensities and often introduce visible artifacts, frequency-domain perturbations (\textit{e.g.} via 2D-DCT) offer a more perceptually aware and structured space. The DCT separates image content by frequency, with low-frequency coefficients encoding global structure and high-frequency ones capturing fine details. This allows us to introduce distortions into selected bands that are more vulnerable to attacks but less perceptible to the human eye. Prior work~\cite{freq_adv1,freq_adv2} has shown that DCT attacks achieve a better trade-off between invisibility and attack effectiveness. By operating in the DCT domain, we improve the robustness of the watermark while preserving visual quality and training stability.
The forward 2D-DCT is computed with the formula given by
\begin{table}[]
\centering
\caption{This table illustrates the impact of varying $\alpha_{adv}^2$ on the \textbf{Identity} value  of the watermarking model.}
\label{tab:alpha_2}
\begin{tabular}{c|cccc}
\toprule
& \multicolumn{4}{c}{$\alpha_{adv}^2$} \\
Methods & 0.02 & 1.0 & 5.0 & 7.5 \\ 
\midrule
$\text{DCT-Trans}$ & \textbf{99.953} & \textbf{99.477} & \textbf{72.517} & \textbf{71.471} \\
Trans & 98.345 & 96.905 & 54.551 & 59.651 \\
\bottomrule
\end{tabular}
\vspace{-15pt}
\end{table}
\begin{equation}
{
\scalebox{0.8}{
    \label{eq:dct}
    \begin{math}
    \begin{aligned}
        \boldsymbol{ X_{i,j}}=&\frac{1}{\sqrt[]{2N}}k(i)k(j) \sum_{x=0}^{N-1} \sum_{y=0}^{N-1} I_{x,y}\cos\left [ \frac{(2x + 1)i\pi }{2N} \right ]\cos\left [ \frac{(2y + 1)j\pi }{2N} \right ],
    \end{aligned}
    \end{math}
    }
}
\end{equation}
where $k(i)$, $k(j)$ are the normalizing factors, $I_{x,y}$ is the pixel value at location $(x,y)$  and $X_{i,j}$ is the DCT coefficient at $(i,j)$. To integrate the DCT with an adversarial attack network, our first step involves refining the watermarked image in the YUV color space, followed by computing the forward 2D-block-DCT on each $8\times8$ non-overlapping block of the watermarked image. Next, we selectively mask the high-frequency components of the DCT coefficients for each YUV channel while preserving the remaining parts to apply perturbations. Then, the intermediate DCT representation of the watermarked image is passed through the Transformer or CNN. Finally, the procedure ends with applying the inverse DCT to obtain the output followed by converting it back to the RGB color space from the YUV color space. Additionally, since we hope to fully utilize the potential of the self-attention mechanism in the Transformer, there is a minor difference in procedure between CNN and the Transformer. Specifically, before passing the intermediate DCT representation through Transformer, we divide the intermediate DCT representation into 8$\times$8 pixel blocks. Afterward, we rearrange them in the order of frequency bands (we put the DCT coefficients of the same frequency band from different blocks together) and then pass them through the Transformer. We demonstrate the detailed procedures for integrating the DCT process into both CNN-based and Transformer-based attack networks in Algorithm~\ref{DCT_watermark_algo}. As illustrated in Table~\ref{tab:analysis_network_coco}(b), the application of DCT markedly enhances the performance of the Transformer-based method. However, the CNN-based attack network shows limited improvement, primarily due to its restricted receptive field, which prevents it from effectively capturing signal correlations across different frequencies.

To avoid naming confusion, in the following sections, ``DCT-CNN'' and ``DCT-Transformer'' refer to the CNN- and Transformer-based attack networks integrated with the DCT process. The original CNN-based and Transformer-based attack networks are referred to as ``CNN'' and `Transformer'', respectively.

%%%%%%%%%%%%%%%%%%%%%%%%%%%%%%%%%%%%%%%%%%%%%%%%%%%%%%%%%%%%%%%%%%%%%%%%%%%%%
\begin{algorithm}[t]
    \caption{Integration of the DCT-Process}
    \label{DCT_watermark_algo}
    \begin{algorithmic}
        \Require
        Watermarked Images $I_{w}$, Attack Network $G_{adv}$;
        \Ensure
        Adversarial Images $I_{adv}$;
        \State Convert color space of $I_{w}$ from RGB to YUV;
        \State Apply 2D-block-DCT on each non-overlapping $8 \times 8$ block of $I_{w}$ to get DCT representation $I_{w}^{DCT}$;
        \State Apply high frequency mask on each block of $I_{w}^{DCT}$;
        \If{$G_{adv} \in \text{Transformer}$}
            \State Split $I_{w}^{DCT}$ into $8 \times 8$ blocks;
            \State Rearranging DCT coefficients by frequency band across blocks;
        \EndIf
        \State $I_{adv}^{DCT} = G_{adv}(I_{w}^{DCT})$;
        \State Apply 2D-IDCT on $I_{adv}^{DCT}$ to get $I_{adv}$;
        \State Convert the color space of $I_{adv}$ from YUV back to RGB;
        \State \Return $I_{adv}$;
    \end{algorithmic}
\end{algorithm}
%%%%%%%%%%%%%%%%%%%%%%%%%%%%%%%%%%%%%%%%%%%%%%%%%%%%%%%%%%%%%%%%%%%%%%%%%%%%%

\begin{figure*}[ht]
  \centering
  \includegraphics[width=1.0\linewidth]{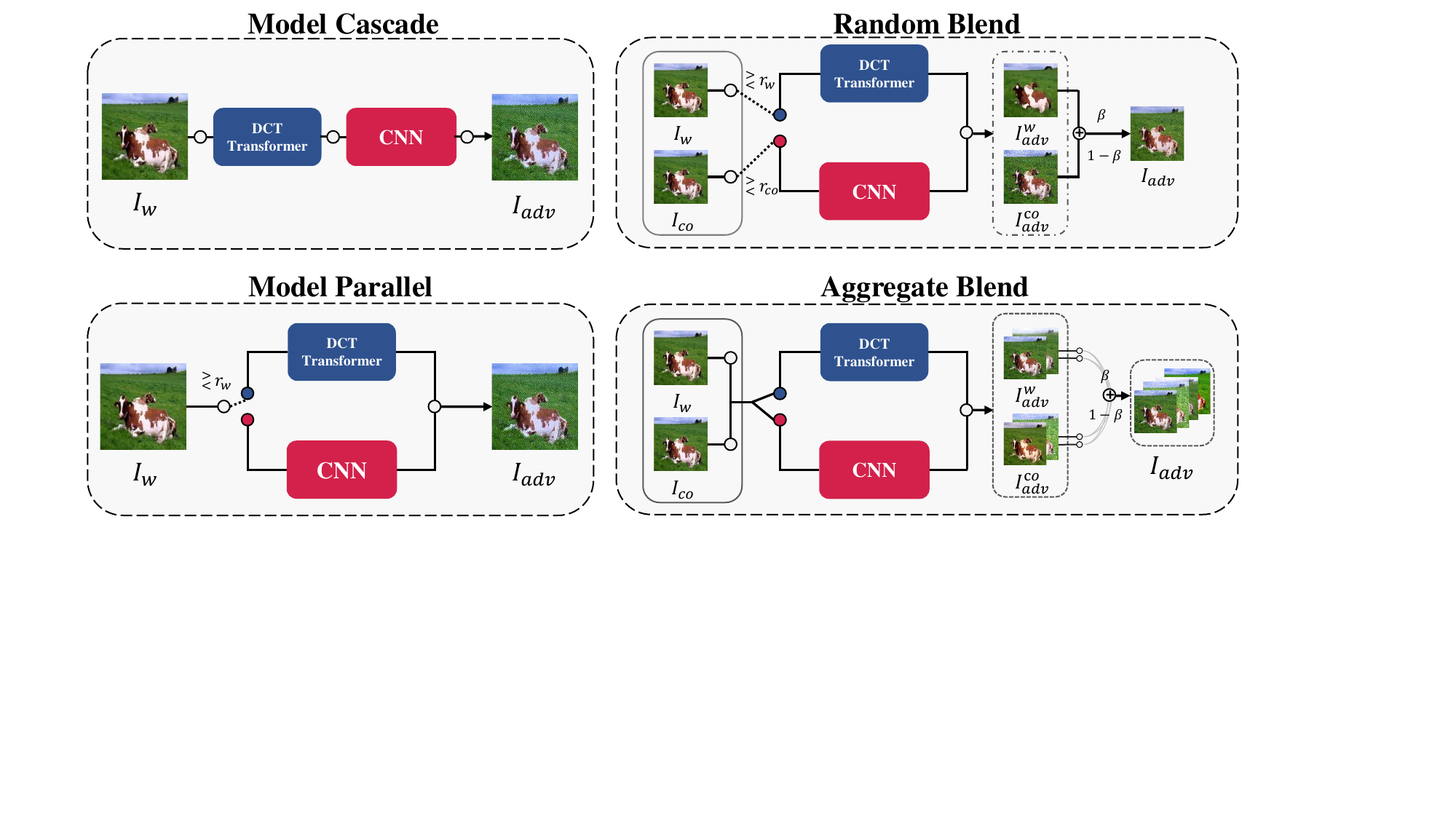}
  \vspace{-115pt}
  \caption{The illustration of four distinct configurations of ensemble methods, including Model Cascade, Model Parallel, Random Blend, and Aggregate Blend. These methods employ various strategies to generate adversarial images $I_{adv}$ from watermarked images $I_{w}$ and cover images $I_{co}$.}
  \label{fig:ensemble}
\end{figure*}
\begin{table*}[ht]
\caption{All evaluation settings remain identical to those outlined in Table~\ref{tab:analysis_network_coco}. This table displays the decoded watermark accuracy for ensemble models, including Model Cascade, Model Parallel, Random Blend, and Aggregate Blend. Each ensemble model consists of different configurations combining the DCT-Transformer with the CNN.}
\label{tab:analysis_network_ensemble}
\scalebox{0.78}{
\begin{tabular}{l|ccccccccccccc|c}
\toprule
Methods & Identity & \begin{tabular}[c]{@{}c@{}}\textcolor{red}{resizedcrop}\\ $p=10 \sim 15$\%\end{tabular} & \begin{tabular}[c]{@{}c@{}}\textcolor{red}{erasing}\\ $p=5 \sim 25\%$\end{tabular} & \begin{tabular}[c]{@{}c@{}}\textcolor{red}{brightness}\\ $p=20 \sim 100\%$\end{tabular} & \begin{tabular}[c]{@{}c@{}}\textcolor{black}{blurring}\\ $p=4 \sim 20 pix$\end{tabular} & \begin{tabular}[c]{@{}c@{}}\textcolor{black}{rotation}\\ $p=9^\circ \sim 45^\circ$\end{tabular} & \begin{tabular}[c]{@{}c@{}}\textcolor{blue}{contrast}\\ $p=20 \sim 100\%$\end{tabular} & \begin{tabular}[c]{@{}c@{}}\textcolor{blue}{noise}\\ $std=0.02 \sim 0.1$\end{tabular} & \begin{tabular}[c]{@{}c@{}}\textcolor{blue}{compression}\\ $p=90 \sim 10$\end{tabular} & Avg \\
\midrule
Model Cascade & 99.947 & 87.573 & 98.452 & 98.026 & 49.859 & 52.338 & 98.268 & 92.898 & 87.236 & 84.955 \\
Model Parallel & 99.851 & 86.28 & 99.364 & 99.448 & 50.378 & 54.178 & 99.024 & 96.464 & 92.997 & \textbf{86.442} \\
Random Blend & 99.976 & 85.896 & 99.807 & 98.437 & 50.092 & 44.992 & 99.324 & 91.174 & 86.377 & 84.008 \\
Aggregate Blend & 99.984 & 85.558 & 99.713 & 98.787 & 49.99 & 48.574 & 99.257 & 87 & 86.634 & 83.944 \\
\bottomrule
\end{tabular}
}
\end{table*}

\subsection{Ensemble Models}
\label{sec:ensemble}
By comparing the performance of watermarking models trained with CNN and DCT-Transformer, as shown in Table~\ref{tab:analysis_network_coco}, we observe that each watermarking model excels under different types of distortions. To enhance the robustness of the watermarking model against diverse distortions, we explore the combination of CNN and DCT-Transformer in four configurations: \textit{Model Cascade, Model Parallel, Aggregate Blend}, and \textit{Random Blend}. The detailed procedures of these four ensemble methods are illustrated in Figure~\ref{fig:ensemble}.

\noindent\textbf{Model Cascade.} During each iteration of the training process, the watermarked image $I_w$ is initially passed through the DCT-Transformer, and then forwarded to CNN for distortion generation.

\noindent\textbf{Model Parallel.} In this dual-path setup, the watermarked image $I_w$ is randomly passed through either the DCT-Transformer or the CNN based on a predetermined threshold $r_w$. For each iteration, a random value in the interval [0,1] is generated. If the value exceeds the threshold $r_w$, the watermarked image $I_w$ will be processed by the DCT-Transformer; otherwise, it is passed through the CNN.

\noindent\textbf{Random Blend.} To preserve the original pixel information, both the cover image $I_{co}$ and the watermarked image $I_w$ are used as inputs to the attack network. For each image, a random threshold $r_{co}$ for $I_{co}$ and $r_w$ for $I_w$ is used to decide whether it will be processed by the DCT-Transformer or the CNN. Importantly, in each iteration, exactly one of the two images is passed through the DCT-Transformer and the other through the CNN, ensuring a complementary processing path. After generating $I_{adv}^{co}$ and $I_{adv}^w$, the two outputs are blended using a blending factor $\beta$ according to the equation $I_{adv} = \beta  \cdot I_{adv}^w + (1-\beta) \cdot I_{adv}^{co} $, where $\beta \in [0,1]$ is randomly generated in each iteration.

\noindent\textbf{Aggregate Blend.} In contrast to the Random Blend method, both the cover image $I_{co}$ and the watermarked image $I_w$ are passed through the DCT-Transformer and the CNN, resulting in two sets of adversarial images, $I_{adv}^{co}$ and $I_{adv}^w$, each containing two adversarial images. We randomly select one image from each sets and blend the selected images together using a blending coefficient $\beta$. Consequently, the final adversarial output $I_{adv}$ is an aggregation of many possible combinations.

\begin{figure*}[t]
\centering
\subfloat[]{\label{fig:multi_level_distortion}
\centering
\scalebox{0.8}{
\includegraphics[width=0.28\linewidth]{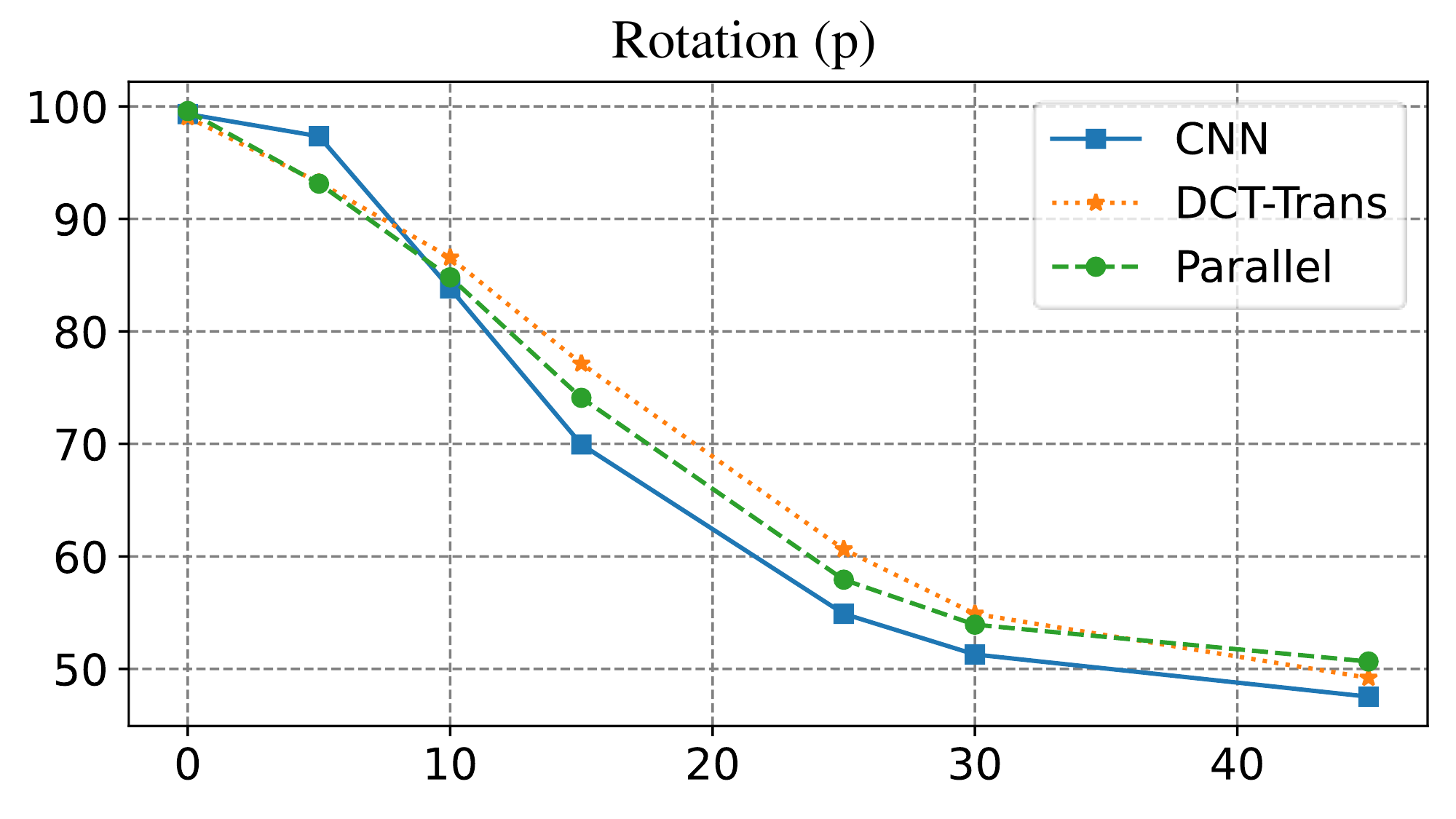}}
}
\subfloat[]{\label{fig:cdiagram}
\centering
\scalebox{0.8}{
\includegraphics[width=0.28\linewidth]{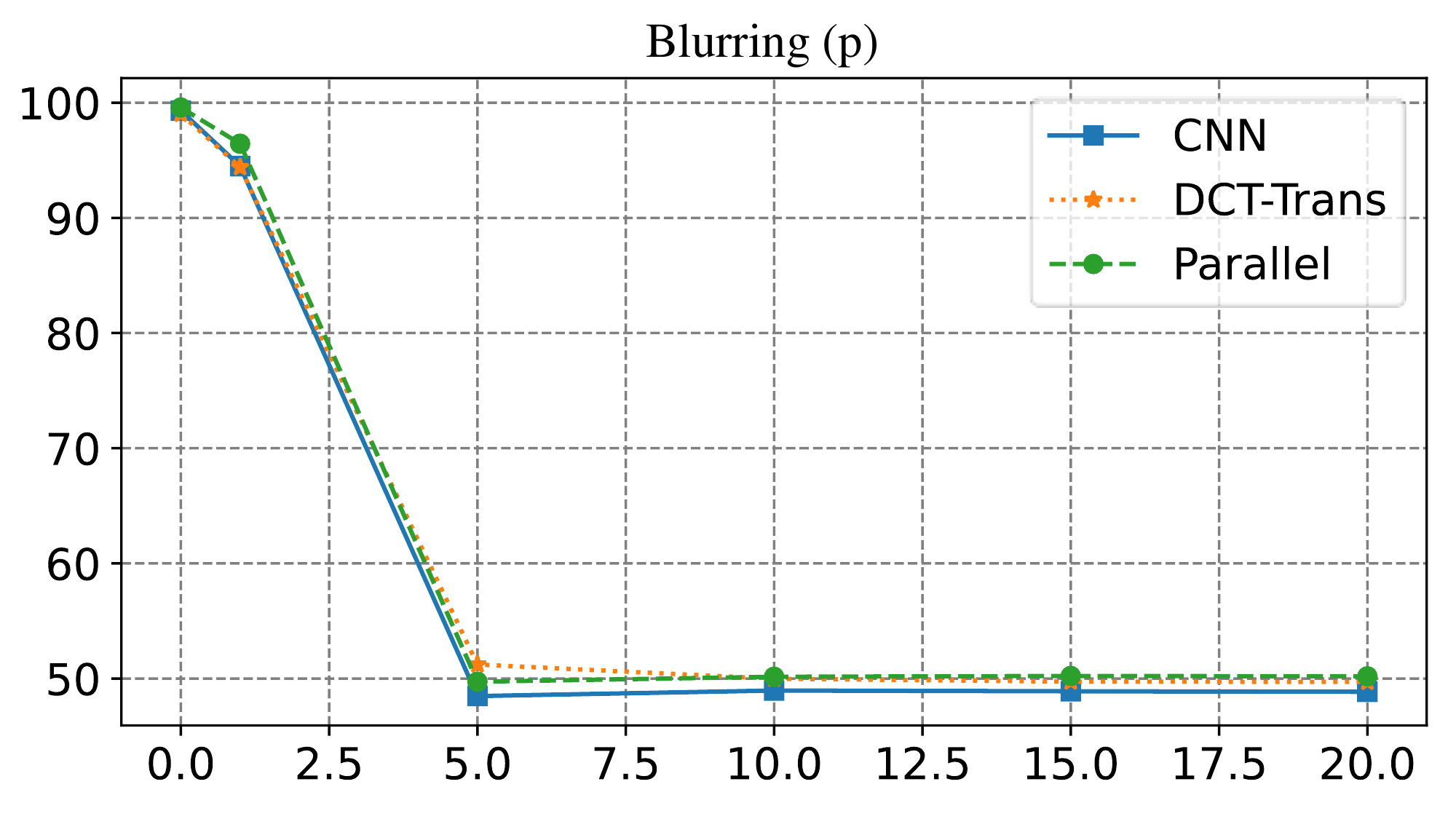}}
}
\label{fig:csetup}
\subfloat[]{\label{fig:cdiagram}
\centering
\scalebox{0.8}{
\includegraphics[width=0.28\linewidth]{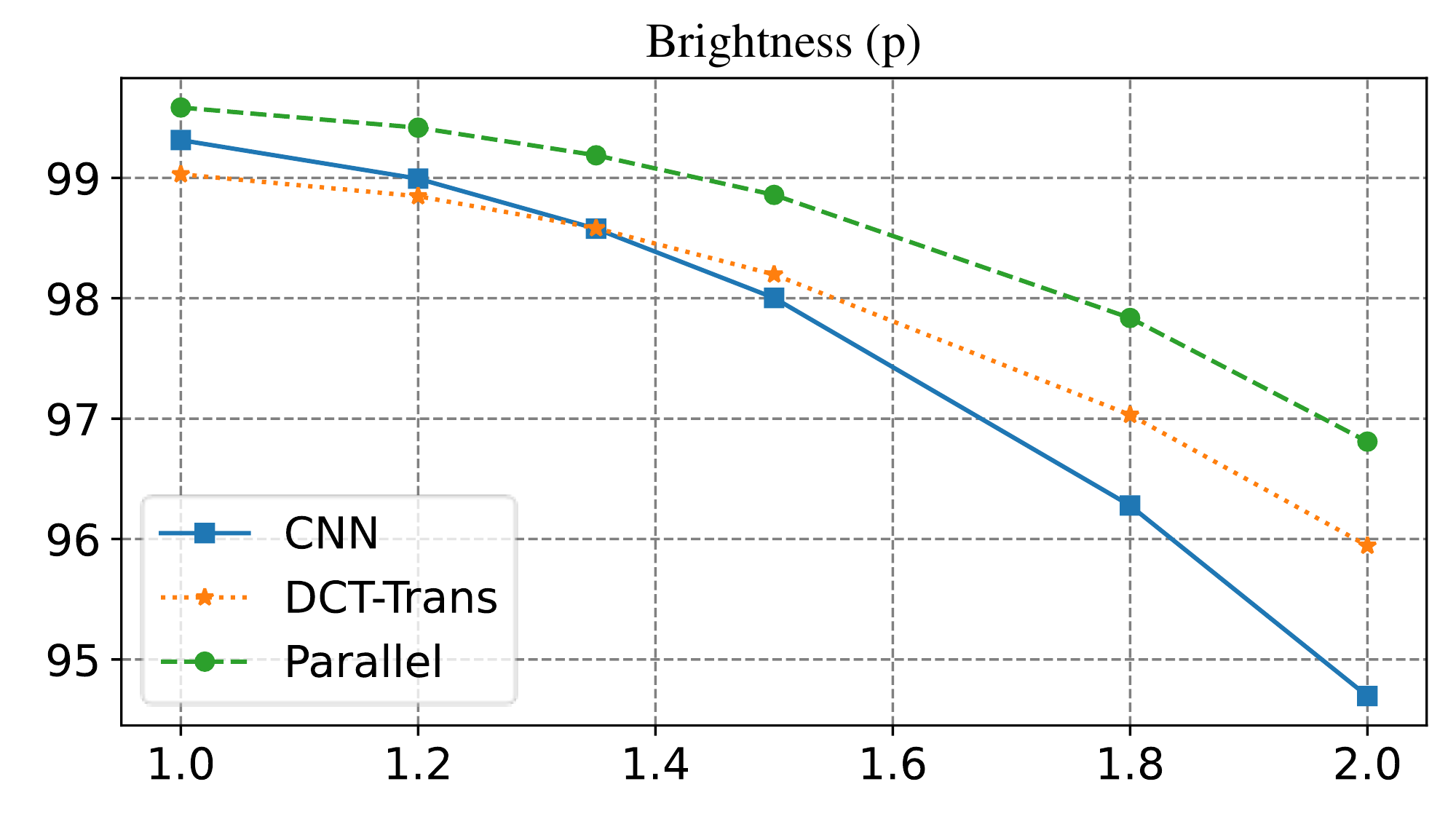}}
}
\label{fig:csetup}
\subfloat[]{\label{fig:cdiagram}
\centering
\scalebox{0.8}{
\includegraphics[width=0.28\linewidth]{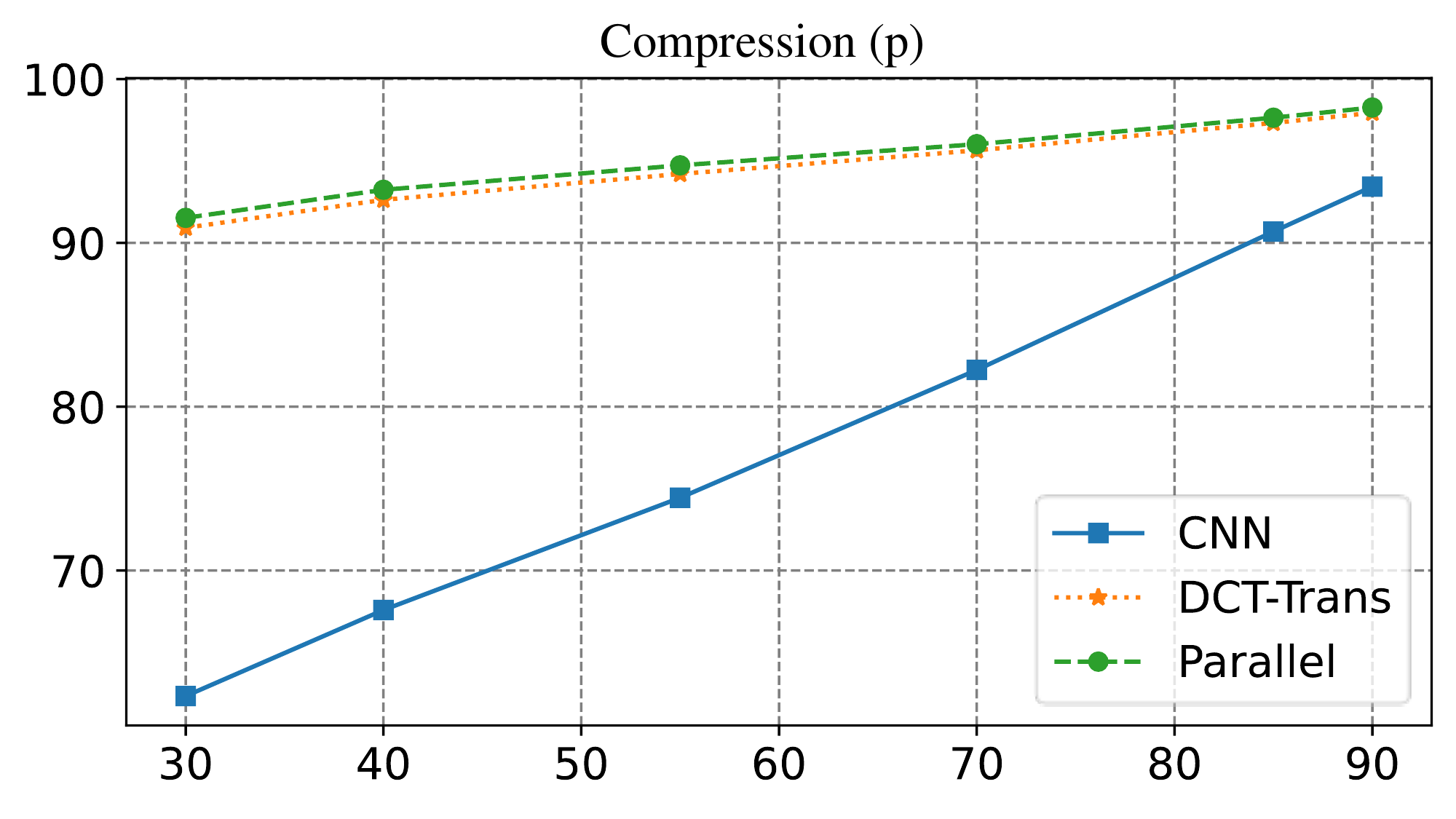}}
}
\label{fig:csetup}
\subfloat[]{\label{fig:cdiagram}
\centering
\scalebox{0.8}{
\includegraphics[width=0.28\linewidth]{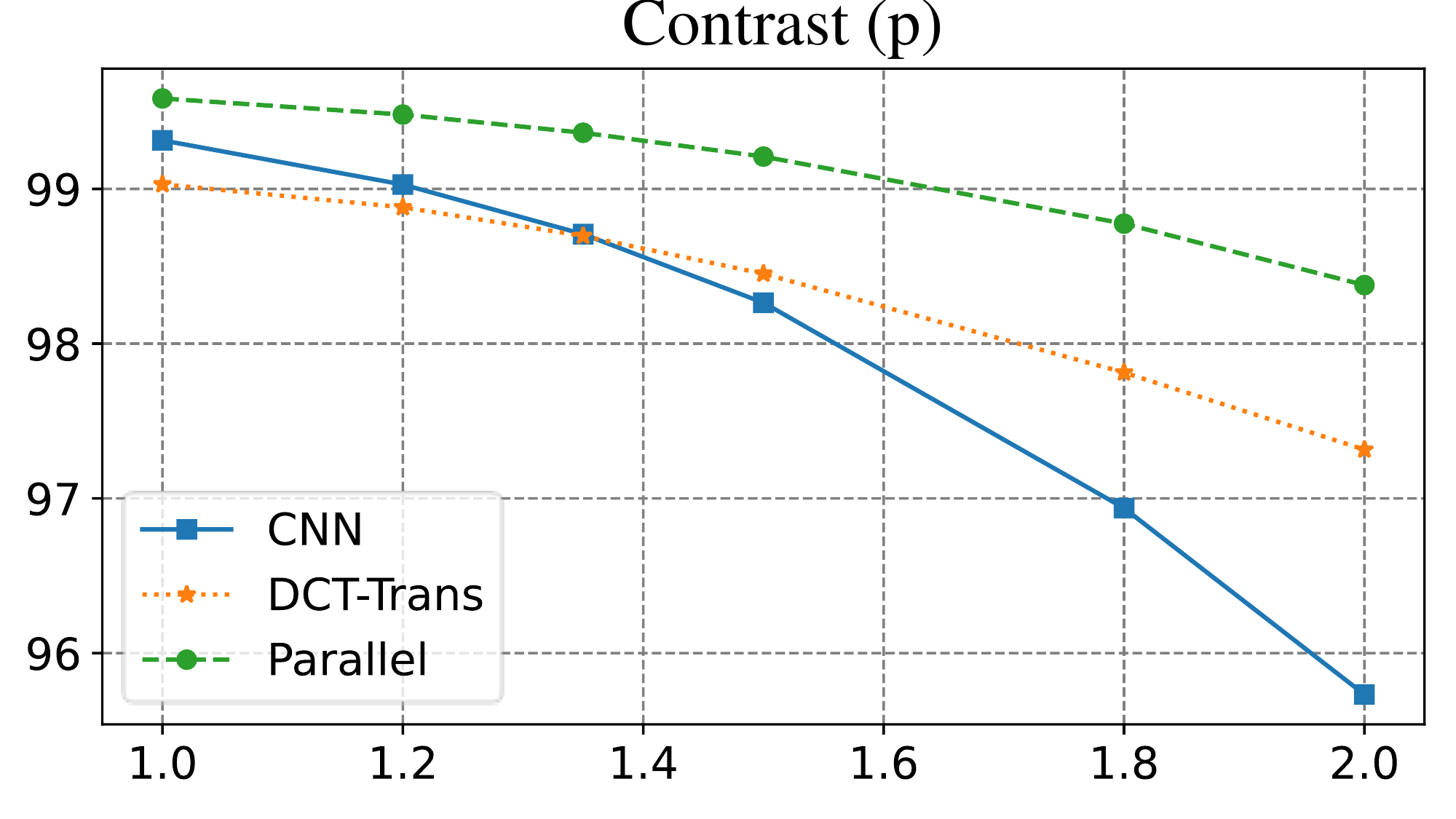}}
}
\label{fig:csetup} 
\subfloat[]{\label{fig:cdiagram}
\centering
\scalebox{0.8}{
\includegraphics[width=0.28\linewidth]{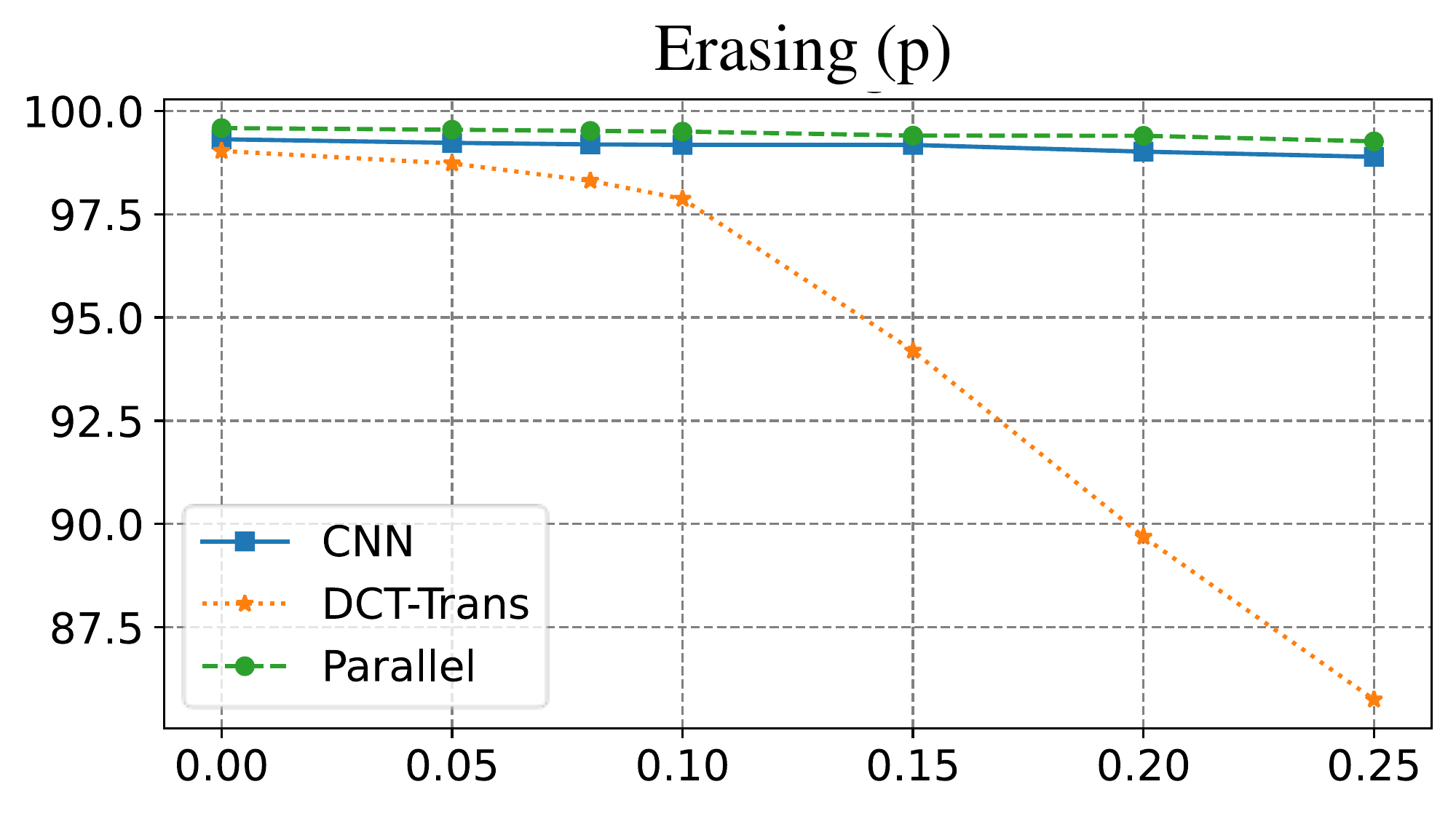}}
}
\label{fig:csetup}
\subfloat[]{\label{fig:cdiagram}
\centering
\scalebox{0.8}{
\includegraphics[width=0.28\linewidth]{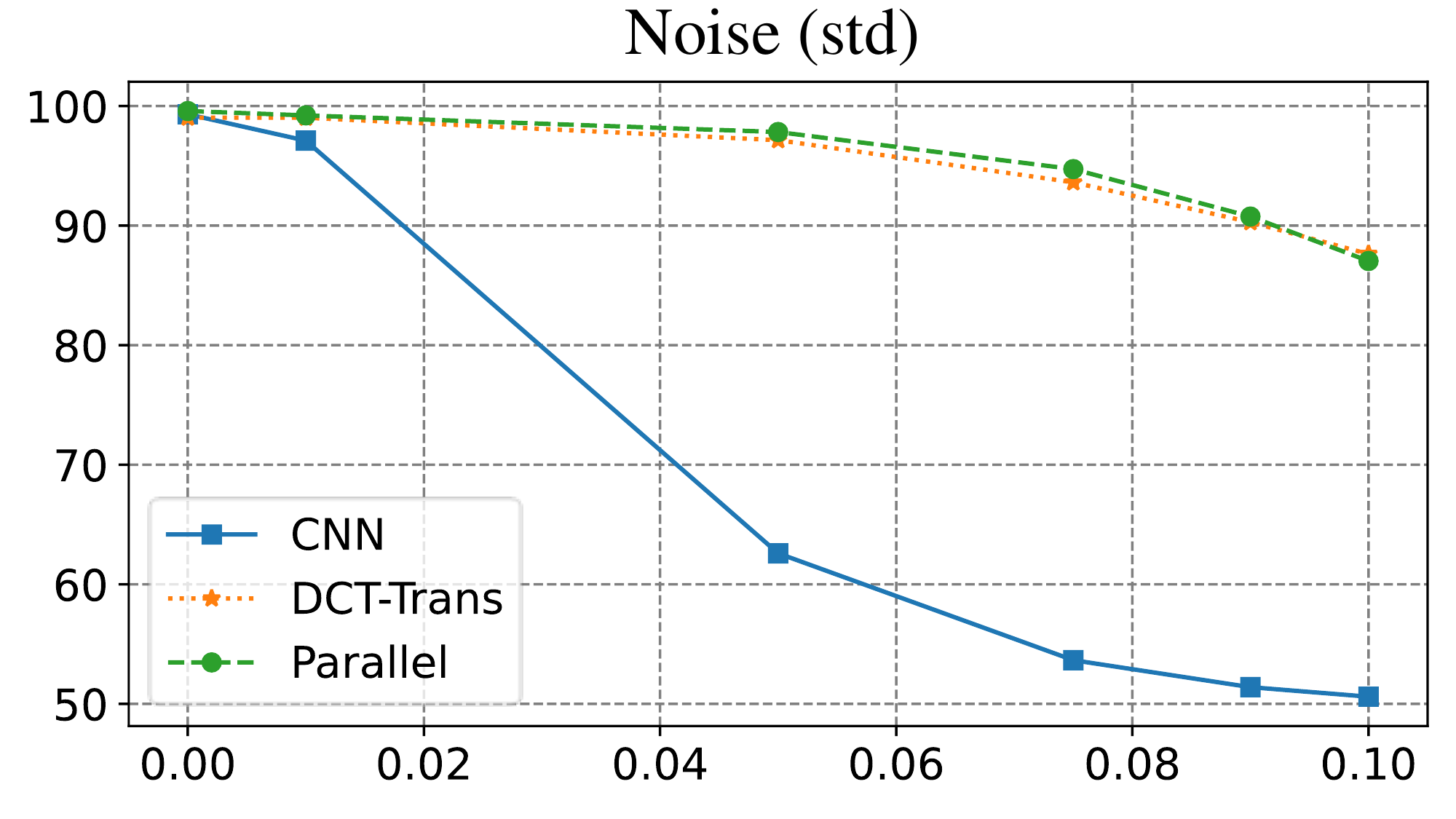}}
}
\label{fig:csetup}
\subfloat[]{\label{fig:cdiagram}
\centering
\scalebox{0.8}{
\includegraphics[width=0.28\linewidth]{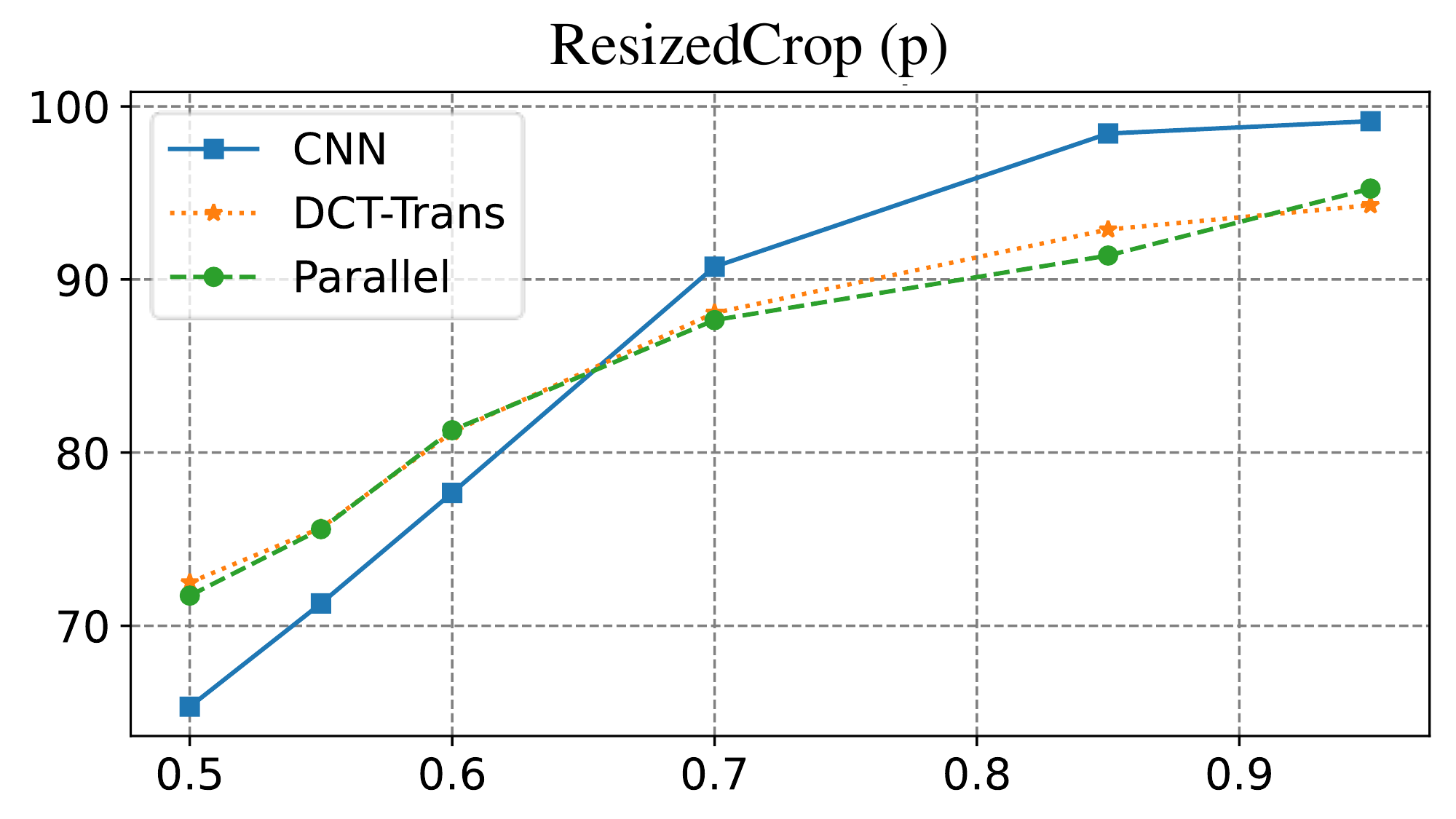}}
}
\label{fig:csetup}
\vspace{-9pt}
\caption{We compare the bit accuracy (y-axis) of different watermarking models trained with distinct attack networks. The proposed ensemble method, which integrates CNN and DCT-Transformer attack networks, achieves performance comparable to models trained with either CNN or DCT-Transformer alone across all distortion parameters (x-axis, shown as $\textit{p}$ or $\textit{std}$ depending on the attack). This demonstrates that our method effectively combines the strengths of both architectures.}
\label{fig:result_range}
\end{figure*}

Among four ensemble strategies, Model Parallel (MP) attains the highest average bit accuracy, as shown in Table~\ref{tab:analysis_network_ensemble}, suggesting that balanced exposure to diverse distortions helps prevent overfitting to any single distortion characteristic and thereby improves overall robustness. This hypothesis is further supported by the observation that the performance of the watermarking model trained with Model Parallel (MP) generally lies between those trained exclusively with CNN or DCT-Transformer attack networks. For instance, it outperforms the CNN in handling compression distortions while remaining comparable to the DCT-Transformer. These findings are further validated by comparisons under various distortion levels, as illustrated in Figure~\ref{fig:result_range}.

\begin{table*}[]
\centering
\caption{All training settings remain identical to those outlined in Table~\ref{tab:analysis_network_coco}. HiDDeN\_MP refers to the use of Model Parallel as an attack network, jointly trained with HiDDeN’s encoder-decoder network, and the same applies to StegaStamp\_MP. Notably, HiDDeN is a re-implemented version in PyTorch, as Zhu~\textit{et al.}~\cite{hidden} only provided a TensorFlow version. }
\label{tab:analysis_network_celeba}
\vspace{-5pt}
\scalebox{0.7}{
\begin{tabular}{lc|ccccccccccc}
\toprule
Methods & Dataset & Identity & \begin{tabular}[c]{@{}c@{}}\textcolor{red}{resizedcrop}\\ $p=10 \sim 15$\%\end{tabular} & \begin{tabular}[c]{@{}c@{}}\textcolor{red}{erasing}\\ $p=5 \sim 25\%$\end{tabular} & \begin{tabular}[c]{@{}c@{}}\textcolor{red}{brightness}\\ $p=20 \sim 100\%$\end{tabular} & \begin{tabular}[c]{@{}c@{}}\textcolor{black}{blurring}\\ $p=4 \sim 20 pix$\end{tabular} & \begin{tabular}[c]{@{}c@{}}\textcolor{black}{rotation}\\ $p=9^\circ \sim 45^\circ$\end{tabular} & \begin{tabular}[c]{@{}c@{}}\textcolor{blue}{contrast}\\ $p=20 \sim 100\%$\end{tabular} & \begin{tabular}[c]{@{}c@{}}\textcolor{blue}{noise}\\ $std=0.02 \sim 0.1$\end{tabular} & \begin{tabular}[c]{@{}c@{}}\textcolor{blue}{compression}\\ $p=90 \sim 10$\end{tabular} & Avg \\
\midrule
HiDDeN~\cite{hidden} & COCO & 99.11 & 84.479 & 98.547 & 96.806 & \textbf{51.314} & \textbf{56.706} & 97.699 & 55.957 & 57.108 & 77.525 \\
DA~\cite{distortion} & COCO & 99.947 & \textbf{90.301} & 99.1 & 97.917 & 49.991 & 52.699 & 97.852 & 58.793 & 68.423 & 79.447 \\
StegaStamp~\cite{stegastamp} & COCO & 99.999 & 48.265 & \textbf{99.564} & 98.521 & 49.897 & 49.994 & 99.012 & 89.359 & 50.866 & 76.164 \\
HiDDeN\_MP (ours) & COCO & 99.851 & 86.28 & 99.364 & \textbf{99.448} & 50.378 & 54.178 & \textbf{99.024} & \textbf{96.464} & 92.997 & \textbf{86.442} \\
StegaStamp\_MP (ours) & COCO & 99.999 & 50.509 & 96.396 & 95.073 & 50.384 & 50.263 & 97.17 & 95.264 & \textbf{98.981} & 81.559 \\
\midrule
HiDDeN & CelebA & 99.092 & 84.833 & 98.231 & 96.212 & \textbf{51.242} & \textbf{57.426} & 96.492 & 52.438 & 55.293 & 76.806 \\
DA & CelebA & 99.99 & \textbf{90.964} & 98.958 & 97.719 & 49.908 & 53.185 & 97.562 & 54.291 & 63.859 & 78.492 \\
StegaStamp & CelebA & 99.99 & 47.842 & \textbf{99.669} & \textbf{98.883} & 49.853 & 50.012 & \textbf{98.897} & 88.784 & 51.052 & 76.109 \\
HiDDeN\_MP (Ours) & CelebA & 99.592 & 86.045 & 98.864 & 97.787 & 50.422 & 53.959 & 98.287 & \textbf{95.451} & 91.878 & \textbf{85.809} \\
StegaStamp\_MP (Ours) & CelebA & 99.999 & 50.341 & 96.381 & 94.541 & 50.203 & 50.008 & 95.511 & 94.47 & \textbf{98.916} & 81.152 \\
\bottomrule
\end{tabular}
}
\end{table*}

\begin{figure}[t]
  \centering
  \includegraphics[width=1.0\linewidth]{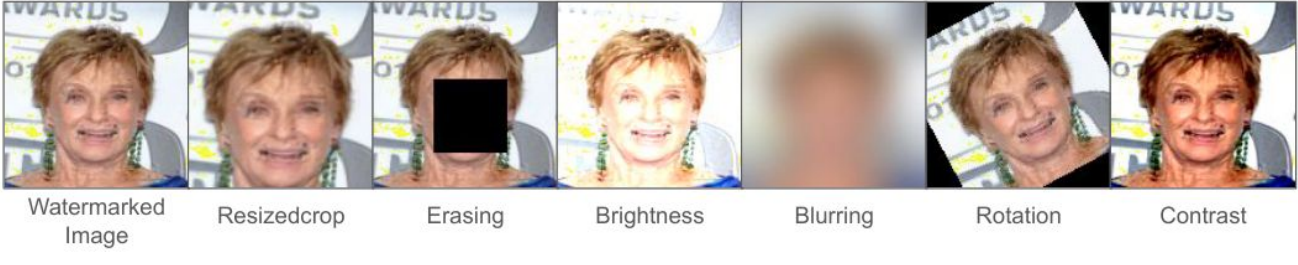}
  \vspace{-20pt}
  \caption{We show the watermarked images after being subjected to the distortions utilized in the evaluation.}
  \label{fig:distorted_images}
\vspace{-15pt}
\end{figure}

\section{Experiments}
\label{sec:exp}
We first provide detailed descriptions of the experimental settings in Section~\ref{sec:exp_details}, and then demonstrate the improvement in watermark robustness achieved by applying the proposed ensemble network to state-of-the-art (SOTA) post-processing watermarking models in Section~\ref{sec:exp_post_processing}. Finally, in Section~\ref{sec:exp_in_processing}, we show that the enhanced post-processing model can also benefit the in-processing watermarking method Stable Signature~\cite{signature}.

\subsection{Experimental Settings}\label{sec:exp_details}
For evaluation, we compute the bit accuracy between encoded and extracted watermarks to assess their robustness in various scenarios. We use the watermark stress tests defined in WAVES~\cite{waves} to evaluate the resistance of watermarks against \textbf{Distortion Attacks}, \textbf{Embedding Attacks}, and \textbf{Regeneration Attacks} in both post-processing and in-processing watermark experiments. In \textbf{Manipulation Attacks}, we evaluate watermark robustness after manipulations made by SOTA image editing models in the post-processing watermark experiment. To ensure both robustness and imperceptibility, we also measure the quality of watermarked images using Peak Signal-to-Noise Ratio (PSNR) and Structural Similarity Index (SSIM).

\noindent\textbf{Implementation Details.}
We train the models on the COCO~\cite{coco} dataset and evaluate their performance on both the COCO and CelebA~\cite{celeba} datasets. Following recent works, we resize the images to $128 \times 128$ pixels for training and evaluation. In the COCO dataset, the training set comprises $110k$ images, and the testing set comprises $40k$ images. For the CelebA dataset, the training set consists of $150k$ images, and the testing set comprises $50k$ images. The channel coding model follows the design used in DA~\cite{distortion}, where the original 30-bit message (watermark) is expanded to 120 bits using the NECST~\cite{necst} model to inject noise redundancy. The Transformer attack network adopts the standard encoder architecture from ViT~\cite{vit}. Specifically, the cover image is split into $8 \times8 $ patches and projected into embeddings with 256 dimensions. The number of encoder layers is $D = 6$, and attention heads $H = 12$ are chosen to optimize the training process. For the Transformer attack network, we set the learning rate to $1e-4$ and the weight decay to $1e-3$. For the CNN attack network, the watermarking components (\textit{e.g.}, Encoder, Decoder, Discriminator), and the loss function hyperparameters, we follow the same settings as DA~\cite{distortion}. The learning rate of both the CNN attack network and the watermarking components is set to $1 \times 10^{-3}$. For the loss function hyperparameters, we set $\alpha_{W_{enc}}^{1} = 1.5$ and $\alpha_{W_{enc}}^{2} = 0.01$ for the watermark encoder, and $\alpha_{W_{dec}}^{1} = 0.3$ and $\alpha_{W_{dec}}^{2} = 0.2$ for the watermark decoder. In the attack network, we set $\alpha_{adv}^{1} = 15.0$ and $\alpha_{adv}^{2} = 1.0$. Finally, in the ensemble methods, both thresholds $r_w$ and $r_{co}$ are set to $0.7$.

\noindent\textbf{WAVES and Manipulation Attacks.} 
To safeguard ownership, watermarks must withstand multiple attacks during transmission or manipulation. Therefore, we utilize the benchmarks defined in WAVES~\cite{waves} to assess the robustness of watermarks. \textbf{Distortion Attacks:} Watermarked images frequently experience distortions during digital transmission. However, most studies evaluate watermark robustness only in isolated or extreme cases. An~\textit{et al.}~\cite{waves} set reasonable-strength distortion attacks as baseline tests for assessing the robustness of watermarks. \textbf{Embedding Attacks:} An~\textit{et al.}~\cite{waves} explore whether adversarial images crafted using the PGD algorithm~\cite{pgd}, within a perturbation limit, can cause off-the-shelf embedding models to deceive the decoder. They utilize pre-trained models such as ResNet18~\cite{resnet18}, CLIP's image encoder~\cite{clip}, KLVAE(f8), SDXL-VAE~\cite{sdxl}, and KL-VAE(f16) in this experiment. \textbf{Regeneration Attacks:} An~\textit{et al.}~\cite{waves} employ diffusion models or VAE~\cite{vaes} to alter an image's latent representation through a process of noising and denoising, to investigate whether watermarked images can resist these processes. Moreover, they explore the robustness of watermarks while undergoing multiple cycles of the noising and denoising process, unlike existing works focus on a single regeneration process. \textbf{Manipulation Attacks:} To test the practicality of using these watermarking models in real-world scenarios, we utilize the Diffusion-based model, InstructPix2Pix~\cite{instructpix2pix}, to transform the style of the watermarked images into a cartoon style. For the GAN-based approach, we employ StyleRes~\cite{styleres} to modify watermarked images based on various prompts, such as changing hair color or altering the depicted individual's gender.

As discussed in Section~\ref{sec:ensemble}, the watermarking model trained with the Model Parallel (MP) attack network achieves the highest average bit accuracy. Therefore, in the following experiments, we compare HiDDeN~\cite{hidden} trained with MP (HiDDeN\_MP) and StegaStamp~\cite{stegastamp} trained with MP (StegaStamp\_MP) against previous methods.

\begin{table}[t]
\caption{Comparison of watermarked image quality between our method and previous approaches.}\label{tab:psnr}
\scalebox{0.8}{
\begin{tabular}{l|lllll|l}
\toprule
\textit{} & \multicolumn{5}{c|}{\textbf{PSNR $\uparrow$}} &  \\ Methods
& \multicolumn{1}{c}{RGB} & \multicolumn{1}{c}{Y} & \multicolumn{1}{c}{U} & \multicolumn{1}{c}{V} & \multicolumn{1}{c|}{B} & SSIM $\uparrow$ \\ \hline
HiDDeN~\cite{hidden} & 33.463 & 34.701 & 40.91 & 37.977 & 33.079 & 0.964 \\
DA~\cite{distortion} & 34.559 & 35.23 & 45.26 & 45.185 & 33.877 & 0.969 \\
StegaStamp~\cite{stegastamp} & \textbf{45.9} & \textbf{50.084} & \textbf{51.309} & \textbf{47.764} & \textbf{42.062} & \textbf{0.995} \\
HiDDeN\_MP (Ours) & 31.062 & 31.193 & 42.968 & 39.343 & 30.813 & 0.947 \\
StegaStamp\_MP (Ours) & 37.623 & 38.747 & 47.168 & 42.999 & 36.142 & 0.982 \\
\bottomrule
\end{tabular}
}
\vspace{-10pt}
\end{table}

\begin{figure}[t]
  \centering
  \includegraphics[width=1.0\linewidth]{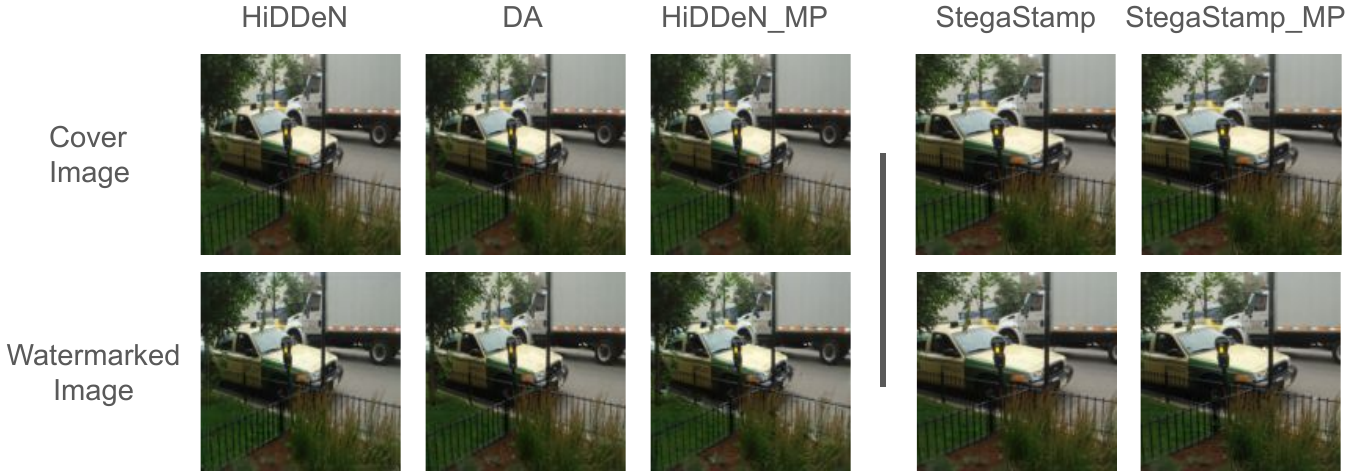}
   \caption{We show the difference between cover images and watermarked images generated by different models.}
   \label{fig:watermark_similar}
\vspace{-15pt}
\end{figure}

\begin{table}[t]
\footnotesize
\centering
\caption{We evaluate previous post-processing watermarking approaches and combine them with our method to test their performance under embedding attacks.}
\label{tab:Embedding_attacks_1}
\scalebox{0.8}{
\begin{tabular}{c|ccccccc}
\toprule
Methods & ResNet18 & CLIP & KLVAE8 & SdxlVAE & KLVAE16 & Avg \\
\hline
HiDDeN~\cite{hidden} & 90.089 & 91.322 & 90.77 & 90.881 & 88.005 & 90.213 \\
DA~\cite{distortion} & 93.931 & 95.057 & 95.534 & 95.153 & 92.58 & 94.451 \\
StegaStamp~\cite{stegastamp} & \textbf{99.876} & 99.319 & \textbf{99.934} & \textbf{99.883} & \textbf{99.874} & \textbf{99.777} \\
HiDDeN\_MP (Ours) & 99.081 & 99.145 & 99.265 & 99.289 & 98.424 & 99.04 \\
StegaStamp\_MP (Ours) & 99.622 & \textbf{99.667} & 99.343 & 99.538 & 97.851 & 99.204 \\
\bottomrule
\end{tabular}
}
\end{table}

\begin{table}[t]
\footnotesize
\centering
\caption{We assess previous post-processing watermarking approaches and integrate them with our method to evaluate their performance against regeneration attacks.}
\label{tab:Regeneration_attacks_1}
\scalebox{0.8}{
\begin{tabular}{c|ccccc}
\toprule
Methods & Regen-Diff & Regen-VAE & Regen-Diff2X & Regen-Diff4X & Avg \\
\hline
HiDDeN~\cite{hidden} & 51.237 & 50.299 & 51.034 & 50.910 & 50.870 \\
DA~\cite{distortion} & 50.599 & 50.228 & 50.408 & 50.388 & 50.405 \\
StegaStamp~\cite{stegastamp} & 49.900 & 50.500 & 50.100 & 50.200 & 50.175 \\
HiDDeN\_MP (Ours) & 56.354 & 50.320 & 55.334 & 53.763 & 53.942 \\
StegaStamp\_MP(Ours) & \textbf{61.396} & \textbf{97.063} & \textbf{59.733} & \textbf{57.48} & \textbf{68.918} \\
\bottomrule
\end{tabular}
}
\vspace{-5pt}
\end{table}

\subsection{Post-processing Watermarks}\label{sec:exp_post_processing}
\textbf{Distortion Attacks.} We train models on COCO training set and embed watermarks in both CelebA and COCO testing sets. The watermarked images are then subjected to a range of distortions to evaluate their resilience. Subsequently, we feed the distorted watermarked images into the decoder to compute the bit accuracy of the extracted watermark. As shown in Table~\ref{tab:analysis_network_celeba}, our method (HiDDeN\_MP and StegaStamp\_MP) outperforms DA~\cite{distortion} in average bit accuracy by 6.995\% on COCO and 7.317\% on CelebA, and improves StegaStamp~\cite{stegastamp} by 5.395\% on COCO and 5.043\% on CelebA. Furthermore, we show distorted images in Figure~\ref{fig:distorted_images} to illustrate that our method can extract watermarks even when strong perturbations have severely degraded the quality of the watermarked image.

Additionally, ensuring both bit accuracy and the perceptual invisibility of watermarks is essential. Although the PSNR and SSIM values of watermarked images produced by the models trained with MP are slightly lower than those of other models (see Table~\ref{tab:psnr}), the images generated by our methods (HiDDeN\_MP and StegaStamp\_MP) still exhibit high visual quality, as shown in Figure~\ref{fig:watermark_similar}.

\noindent\textbf{Embedding Attacks.} We calculate the bit accuracy between the encoded and extracted watermarks after attacks using five off-the-shelf embedding models. We show the experimental results in Table~\ref{tab:Embedding_attacks_1} and demonstrate that the watermark encoder-decoder jointly trained with MP achieves nearly 100\% bit accuracy.

\noindent\textbf{Regeneration Attacks.} We evaluate the robustness of watermarks through noising and denoising process. Watermarked images are passed through DMs or VAEs~\cite{vaes}, where they are mapped to latent representations and subsequently reconstructed. After regeneration, the images are fed into the decoder to extract the watermarks, and the bit accuracy is computed between encoded and decoded watermarks. As shown in Table~\ref{tab:Regeneration_attacks_1}, our method outperforms average bit accuracy by 3.537\% over DA and by 18.743\% over StegaStamp.

\begin{figure}[t]
  \centering
  \includegraphics[width=\linewidth]{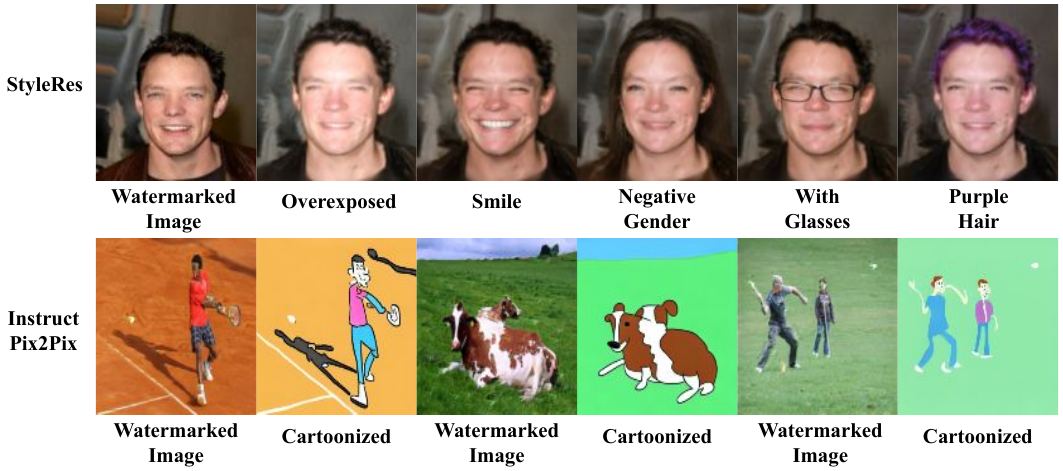}
  \caption{We utilize StyleRes~\cite{styleres} for editing the watermarked images embedded in CelebA with various prompts, as demonstrated in the top row. The second row showcases the outcomes of watermarked images (COCO dataset) after being edited by InstructPix2Pix~\cite{instructpix2pix}, where we employ the prompts ``cartoon, doodle'' for editing the images.}
  \label{fig:editing_images}
\vspace{-15pt}
\end{figure}

\noindent\textbf{Manipulation Attacks.} To assess the robustness of watermarks against image editing models, we utilize SOTA GAN-based and diffusion-based models, StyleRes~\cite{styleres} and InstructPix2Pix~\cite{instructpix2pix}, to manipulate the watermarked images using various prompts. Specifically, we embed watermarks into the CelebA testing set, which contains 50k images, and apply StyleRes to edit the images with attribute-based prompts. For example, altering gender or adding a smile. Similarly, we embed watermarks into the COCO testing set, comprising 40k images, and use InstructPix2Pix to apply edits with the prompts ``Cartoon and Doodle.'' Examples of the edited images are shown in Figure~\ref{fig:editing_images}. We then input the edited images into the decoder to evaluate its ability to recover the embedded watermarks. As shown in Table~\ref{tab:Editing_results}, our method outperforms average bit accuracy by 9.963\% over DA~\cite{distortion} and by 7.551\% over StegaStamp~\cite{stegastamp}.
\begin{table*}[t]
\centering
% \vspace{-15pt}
\caption{All methods are trained on the COCO dataset. After editing by generative models, we compute the bit accuracy of watermarked images in both the COCO and CelebA datasets. In the COCO dataset, we utilize InstructPix2Pix to edit the images. In the CelebA dataset, we employ StyleRes for image editing.}
\label{tab:Editing_results}
\vspace{-5pt}
\scalebox{0.9}{
\begin{tabular}{l|c|ccccc|c}
\toprule
& COCO (InstructPix2Pix) & \multicolumn{6}{c}{CelebA (StyleRes)} \\
\midrule
Methods & Cartoon, Doodle & Smile  & Glasses & Negative Gender & Overexposed & Purple Hair & Avg \\
\midrule
HiDDeN~\cite{hidden} & 59.841 & 51.299 & 51.58 & 50.993 & 51.355 & 51.127 & 52.699 \\
DA~\cite{distortion} & 69.132 & 59.121 & 59.762 & 56.108 & 60.087 & 59.993 & 60.700 \\
Stegastamp~\cite{stegastamp} & 52.142 & 50.040 & 49.957 & 49.948 & 49.992 & 50.004 & 50.347 \\
HiDDeN\_MP (Ours) & 70.984 & \textbf{70.780} & \textbf{72.435} & \textbf{63.477} & \textbf{73.999} & \textbf{72.306} & \textbf{70.663} \\
StegaStamp\_MP (Ours)& \textbf{77.731} & 53.161 & 54.348 & 53.693 & 54.259 & 54.200 & 57.898 \\
\bottomrule
\end{tabular}
}
\end{table*}
\begin{table*}[ht]
\centering
\vspace{-5pt}
\caption{``Stable Signature MP'' refers to fine-tuning VAE decoder using the enhanced watermark decoder trained with MP. We follow the training procedure outlined in Stable Signature~\cite{signature} to fine-tune the VAE. Notably, the performance of Stable Signature is reproduced using a HiDDeN decoder trained by us, and the evaluation is carried out on the COCO dataset.}
\label{tab:exp_in_processing_distortion}
\vspace{-5pt}
\scalebox{0.7}{
\begin{tabular}{l|ccccccccccccc|c}
\toprule
Methods & Identity & \begin{tabular}[c]{@{}c@{}}\textcolor{black}{resizedcrop}\\ $p=10 \sim 15$\%\end{tabular} & \begin{tabular}[c]{@{}c@{}}\textcolor{black}{erasing}\\ $p=5 \sim 25\%$\end{tabular} & \begin{tabular}[c]{@{}c@{}}\textcolor{black}{brightness}\\ $p=20 \sim 100\%$\end{tabular} & \begin{tabular}[c]{@{}c@{}}\textcolor{black}{blurring}\\ $p=4 \sim 20 pix$\end{tabular} & \begin{tabular}[c]{@{}c@{}}\textcolor{black}{rotation}\\ $p=9^\circ \sim 45^\circ$\end{tabular} & \begin{tabular}[c]{@{}c@{}}\textcolor{black}{contrast}\\ $p=20 \sim 100\%$\end{tabular} & \begin{tabular}[c]{@{}c@{}}\textcolor{black}{noise}\\ $std=0.02 \sim 0.1$\end{tabular} & \begin{tabular}[c]{@{}c@{}}\textcolor{black}{compression}\\ $p=90 \sim 10$\end{tabular} & Avg \\
\midrule
Stable Signature~\cite{signature} & 99.863 & 66.264 & 71.653 & 72.467 & \textbf{46.341} & \textbf{49.782} & 75.166 & 64.625 & 58.18 & 67.149 \\
Stable Signature MP (Ours)& 99.741 & \textbf{80.241} & \textbf{88.829} & \textbf{87.039} & 38.247 & 48.109 & \textbf{89.53} & \textbf{82.94} & \textbf{65.143} & \textbf{75.535} \\
\bottomrule
\end{tabular}
}
\end{table*}

\subsection{In-processing Watermarks}\label{sec:exp_in_processing}
Fernandez~\textit{et al.}~\cite{signature} utilize a pre-trained HiDDeN decoder to fine-tune the autoencoder of latent diffusion model (LDM)~\cite{ldm} with a predefined watermark, guiding LDM to generate images that inherently contain the watermark. This approach allows for simultaneous image and watermark generation in a single step.

Building on the effectiveness of our proposed ensemble attack network (MP) in improving the robustness of post-processing watermarking models, we extend our evaluation to in-processing watermarking. Specifically, we fine-tune LDM using both the original HiDDeN decoder and the decoder trained with MP. After image generation, we compute the bit accuracy between the original encoded watermark and the extracted one. Notably, we use the HiDDeN model that we trained to conduct the experiments.

\noindent\textbf{Distortion Attacks.} We apply a series of distortions defined in WAVES~\cite{waves} to attack the watermarked images and evaluate the bit accuracy of the degraded outputs. As shown in Table~\ref{tab:exp_in_processing_distortion}, LDM fine-tuned with decoder trained with MP outperforms the one fine-tuned with original HiDDeN decoder by 8.386\% in average bit accuracy.

\noindent\textbf{Embedding Attacks.} We utilize five off-the-shelf embedding models to attack watermarked images generated by LDM and calculate the bit accuracy between the encoded and extracted watermarks. As shown in Table~\ref{tab:Embedding_attacks_2}, the average bit accuracy achieves 90.078\% by applying MP.

\noindent\textbf{Regeneration Attacks.} We send the generated image as input to diffusion models or VAE, mapping it to the latent space and reconstructing it. After reconstruction, we input the image into the decoder to extract the watermarks and calculate the bit accuracy between the encoded and decoded watermarks. As shown in Table~\ref{tab:Regeneration_attacks_2}, our method is slightly lower than Stable Signature by 0.605\% while still being comparable. \\
\vspace{-10pt}

\begin{table}[t]
\footnotesize
\centering
\caption{We compare the performance of Stable Signature and our method under embedding attacks, with all evaluations performed on the COCO dataset.}
\label{tab:Embedding_attacks_2}
\scalebox{0.8}{
\begin{tabular}{c|ccccccc}
\toprule
Methods & ResNet18 & CLIP & KLVAE8 & SdxlVAE & KLVAE16 & Avg \\
\hline
Stable Signature~\cite{signature} & 89.307 & 89.335 & 89.721 & 89.446 & 88.845 & 89.330 \\
Stable Signature MP (Ours) & \textbf{90.235} & \textbf{90.098} & \textbf{90.396} & \textbf{89.966} & \textbf{89.695} & \textbf{90.078} \\
\bottomrule
\end{tabular}
}
\vspace{-10pt}
\end{table}

\begin{table}[t]
\footnotesize
\centering
\caption{We compare the performance of Stable Signature and our method under regeneration attacks, with all evaluations performed on the COCO dataset.}
\label{tab:Regeneration_attacks_2}
\scalebox{0.8}{
\begin{tabular}{c|cccccc}
\toprule
Methods & Regen-Diff & Regen-VAE & Regen-Diff2X & Regen-Diff4X & Avg \\
\hline
Stable Signature~\cite{signature} & \textbf{50.926} & 56.829 & \textbf{52.617} & \textbf{54.383} & \textbf{53.688} \\
Stable Signature MP (Ours) & 49.536 & \textbf{61.365} & 50.390 & 51.042 & 53.083
\\
\bottomrule
\end{tabular}
}
\end{table}

% \begin{figure}[t]
%   \centering
%   \includegraphics[width=\linewidth]{Imgs/adv_imgs.pdf}
%   \caption{Among the adversarial images generated by various attack networks, the DCT-Transformer results in the most severe image degradation.}
%   \label{fig:adv_imgs}
% \end{figure}
\begin{table}[t]
\caption{Results of varying the number of transformer encoder blocks (Depth) and attention heads (Head) in the DCT-Transformer attack network. $\text{AVG}_{\text{ALL}}$ denotes the average bit accuracy across all distortions that are shown in Table~\ref{tab:analysis_network_coco}.}
\label{tab:ablation_depth}
\scalebox{0.9}{
\begin{tabular}{ccccccc}
\toprule
\multicolumn{7}{c}{Transformer Configurations}                   \\ \hline
Depth & Head & params & \begin{tabular}[c]{@{}c@{}}Identity\end{tabular} & \begin{tabular}[c]{@{}c@{}}Dropout\\ $p=0.3$\end{tabular} &   \begin{tabular}[c]{@{}c@{}}Sat\\ $f=15.0$\end{tabular} &
  \begin{tabular}[c]{@{}c@{}} $\text{AVG}_{\text{ALL}}$ \end{tabular} \\ \hline
  6  & 12 & 5.664M  & 99.678 & \underline{90.103}  & \textbf{84.702} & \textbf{85.914}\\
12 & 12 & 11.180M   & \underline{99.837} & 89.946 &  \underline{84.699} & 79.706 \\
24 & 12 & 22.211M   & \textbf{99.958} & \textbf{94.405}  & 83.648 &  \underline{79.832}\\ \bottomrule
\end{tabular}
}
\end{table}

\begin{table}[t]
\caption{Using the same settings as Table~\ref{tab:ablation_depth}, this table shows the results of the DCT-Transformer attack network with and without positional embedding.}
\label{tab:ablation_pos_emb}
\scalebox{0.9}{
\begin{tabular}{ccccccc}
\toprule
\multicolumn{7}{c}{Transformer Configurations}           \\ \hline
Depth & Head & + Pos emb & \begin{tabular}[c]{@{}c@{}}Identity\end{tabular} & \begin{tabular}[c]{@{}c@{}}Dropout\\ $p=0.3$\end{tabular} &   \begin{tabular}[c]{@{}c@{}}Sat\\ $f=15.0$\end{tabular} &
  \begin{tabular}[c]{@{}c@{}} $\text{AVG}_{\text{ALL}}$ \end{tabular} \\ \hline
  6 & 12 &          & \underline{99.678} & \textbf{90.103} & \textbf{84.702} & \textbf{85.914} \\
6 & 12 & $\boldsymbol{\checkmark}$ & \textbf{99.768} & \underline{89.031} & \underline{78.565} & \underline{80.046} \\ 
\bottomrule
\end{tabular}
}
\vspace{-5pt}
\end{table}

\begin{table}[t]
\caption{Using the same settings as Table~\ref{tab:ablation_depth}, this table presents the results of applying different color spaces in the DCT-Transformer attack network.}
\label{tab:ablation_color_space}
\scalebox{0.9}{
\begin{tabular}{ccccccc}
\toprule
\multicolumn{7}{c}{Transformer Configurations}           \\ \hline
Depth & Head & Color Space & \begin{tabular}[c]{@{}c@{}}Identity\end{tabular} & \begin{tabular}[c]{@{}c@{}}Dropout\\ $p=0.3$\end{tabular} &   \begin{tabular}[c]{@{}c@{}}Sat\\ $f=15.0$\end{tabular} &
  \begin{tabular}[c]{@{}c@{}} $\text{AVG}_{\text{ALL}}$ \end{tabular} \\ \hline
 
6 & 12 & YUV & \textbf{99.678} & \textbf{90.103} & \textbf{84.702} & \textbf{85.914}  \\
6 & 12 & RGB & \underline{69.530} & \underline{55.351} & \underline{63.520} & \underline{57.935} \\
\bottomrule
\end{tabular}
}
\vspace{-10pt}
\end{table}

\section{Ablation Studies}
\label{sec:ablation_study}
%In Section~\ref{sec:diff_setting},

We conduct extensive ablation studies on the COCO~\cite{coco} dataset to thoroughly explore the synergy between the DCT process and the Transformer architecture. Our investigation examines the impact of various configuration settings within the DCT-Transformer, including the number of standard Transformer encoder blocks (depth), the number of attention heads, the use of positional embeddings, and the choice of color space (YUV versus RGB). %In Section~\ref{sec:visual_adv}, we present visual examples of adversarial images generated by different attack networks. As shown in Figure~\ref{fig:adv_imgs}, these visualizations highlight the distortion generation capabilities of each attack network.

%\subsection{Different Settings of Attack Network}
\label{sec:diff_setting}
\noindent \textbf{Depth of Transformer Block.}
Our analysis focuses on the architecture of the Transformer-based attack network. As shown in Table~\ref{tab:ablation_depth}, increasing the depth of the Transformer by adding more encoder layers does not necessarily improve performance. This is because excessive model complexity can hinder training and increase the risk of overfitting, especially when the added depth does not meaningfully enhance the model’s ability to capture relevant information in the watermarking context.

\noindent \textbf{Positional Embeddings in Transformer.}
Positional embeddings in Transformers typically serve to provide spatial context. However, in the context of our DCT-Transformer, we find that removing positional embeddings improves performance, as shown in Table~\ref{tab:ablation_pos_emb}. This improvement can be attributed to the nature of watermarking and attack processes, where absolute spatial positioning is less critical. Instead, capturing and manipulating frequency-based features, which are more global and resilient to common image transformations, proves to be more effective.

\noindent \textbf{Color Space of DCT-Transformer.}
Our study also investigates the effectiveness of different color spaces, specifically evaluating the performance of the DCT process within these spaces, as shown in Table~\ref{tab:ablation_color_space}. The results demonstrate that the YUV color space outperforms RGB color space. This advantage is likely due to YUV’s separation of luminance (Y) and chrominance (U and V) components. The Y component, which captures brightness information, tends to retain more relevant features for watermarking and is less susceptible to distortions than the chrominance channels. By leveraging this property, applying the DCT process in the YUV color space enhances both robustness and effectiveness of the watermarking model against attacks.

% \subsection{Visualization of Adversarial Images}
% \label{sec:visual_adv}
% As detailed in Section~\ref{sec:dct_vs_spatial}, the integration of the DCT process with attack networks yields significant advantages for the watermarking model's training convergence, particularly when faced with strong perturbations simulated by attack networks. Illustrated in Figure~\ref{fig:adv_imgs}, the integration with the DCT process allows the attack network (DCT-CNN and DCT-Transformer) to generate stronger perturbations compared to those in the spatial domain while maintaining a 99\% Identity value, as depicted in Table~\ref{tab:analysis_network_coco}. Notably, the DCT-Transformer demonstrates the strongest distortion capability among all attack networks, highlighting its effectiveness in significantly enhancing watermark robustness.

\section{Conclusions}
\label{sec:conclusion}
We conduct a comprehensive investigation into the impact of different attack networks and ensemble strategies on watermark robustness, and propose a novel ensemble attack network to enhance the resilience of deep image watermarking pipelines. To evaluate the effectiveness of our proposed approach, we assess watermark robustness using the challenging WAVES benchmark. First, we evaluate the resilience of watermarks under various image distortions. Second, we assess robustness against embedding attacks. Third, we examine performance under noising and denoising processes. Finally, we apply state-of-the-art generative models to manipulate the watermarked images and evaluate whether the decoder can reliably extract the embedded watermarks. Experimental results show that our proposed method consistently enhances the robustness of existing post-processing watermarking models across diverse attack scenarios. Furthermore, we apply the enhanced post-processing model to strengthen the robustness of the in-processing watermarking method.

\noindent\textbf{Limitations.} Since our method employs both Transformer and CNN attack networks to simulate diverse distortions, the computational cost is higher than that of previous approaches that rely solely on CNNs or noise layers. Moreover, although the artifacts in watermarked images generated by our model are not very noticeable, the perceptual metrics such as PSNR and SSIM remain relatively low, leaving room for improvement.

\begin{acks}
    This research is supported by National Science and Technology Council, Taiwan (R.O.C), under the grant number of NSTC-114-2634-F-001-001-MBK, NSTC-113-2634-F-002-007, and NSTC-112-2222-E-001-001-MY2 and Academia Sinica under the grant number of AS-CDA-110-M09 and AS-IAIA-114-M08. We thank to National Center for High-performance Computing (NCHC) of National Institutes of Applied Research (NIAR) in Taiwan for providing computational and storage resources.
\end{acks}


% \begin{acks}
% \end{acks}

%\bibliographystyle{ACM-Reference-Format}
%\bibliography{sample-base}

%\appendix

\end{document}